\documentclass[10pt]{article} 
\usepackage[accepted]{tmlr}
\usepackage{hyperref}
\usepackage{url}

\usepackage{graphicx}
\usepackage{xr}
\usepackage{xcolor}
\usepackage{algorithm}
\usepackage{algpseudocode}
\usepackage{amsthm}
\usepackage{lineno}
\usepackage[utf8]{inputenc}
\usepackage[parfill]{parskip}
\usepackage{color,colortbl}
\definecolor{mydarkblue}{rgb}{0,0.08,0.45}

\usepackage{thmtools, thm-restate}
\usepackage{mdframed}
\usepackage{tikz}
\usepackage{caption}
\usetikzlibrary{decorations.pathreplacing} 

\newmdtheoremenv{mddef}{Definition}

\usepackage{amsmath,amsfonts,bm}
\usepackage{array} 
\usepackage{booktabs} 

\def\eqref#1{equation~\ref{#1}}

\def\1{\bm{1}}

\def\defas{\stackrel{\text{def}}{=}}
\def\EE{\mathbb E}
\def\dif{\mathop{}\!\mathrm{d}}

\newtheorem{definition}{Definition}

\DeclareMathAlphabet{\mathsfit}{\encodingdefault}{\sfdefault}{m}{sl}
\SetMathAlphabet{\mathsfit}{bold}{\encodingdefault}{\sfdefault}{bx}{n}

\usepackage{url}
\usepackage[labelfont=bf]{caption}
\usepackage{soul}

\newcommand{\addblue}[1]{{#1}}

\newcommand{\add}[1]{{#1}}
\newcommand{\remove}[1]{}
\newcommand{\addtwo}[1]{{#1}}
\newcommand{\removetwo}[1]{}

\usepackage[subtle]{savetrees}

\title{Stability-Aware Training of Machine Learning Force Fields with Differentiable Boltzmann Estimators}

\author{\name Sanjeev Raja \email sanjeevr@berkeley.edu \\
      \addr Department of Computer Science, UC Berkeley
      \AND
      \name Ishan Amin \email ishanthewizard@berkeley.edu \\
      \addr Department of Computer Science, Department of Physics, UC Berkeley 
      \AND
      \name Fabian Pedregosa \email pedregosa@google.com \\
      \addr Google Deepmind
      \AND 
      \name Aditi Krishnapriyan \email aditik1@berkeley.edu
       \addr \mbox{Department of Computer Science and Department of Chemical Engineering, UC Berkeley; LBNL}
}

\def\month{02}  
\def\year{2025} 
\def\openreview{\url{https://openreview.net/forum?id=ZckLMG00sO}} 

\begin{document}

\maketitle

\begin{abstract}
Machine learning force fields (MLFFs) are an attractive alternative to \textit{ab-initio} methods for molecular dynamics (MD) simulations. However, they can produce unstable simulations, limiting their ability to model phenomena occurring over longer timescales and compromising the quality of estimated observables. To address these challenges, we present Stability-Aware Boltzmann Estimator (StABlE) Training, a multi-modal training procedure which leverages joint supervision from reference quantum-mechanical calculations and system observables. StABlE Training iteratively runs many MD simulations in parallel to seek out unstable regions, and corrects the instabilities via supervision with a reference observable. We achieve efficient end-to-end automatic differentiation through MD simulations using our Boltzmann Estimator, a generalization of implicit differentiation techniques to a broader class of stochastic algorithms. Unlike existing techniques based on active learning, our approach requires no additional \textit{ab-initio} energy and forces calculations to correct instabilities. We demonstrate our methodology across organic molecules, tetrapeptides, and condensed phase systems, using three modern MLFF architectures. StABlE-trained models achieve significant improvements in simulation stability, data efficiency, and agreement with reference observables. Crucially, the stability improvements cannot be matched by simply reducing the simulation timestep, meaning that StABlE Training effectively allows for larger timesteps in MD simulations. By incorporating observables into the training process alongside first-principles calculations, StABlE Training can be viewed as a general semi-empirical framework applicable across MLFF architectures and systems. This makes it a powerful tool for training stable and accurate MLFFs, particularly in the absence of large reference datasets. Our code is publicly available at \url{https://github.com/ASK-Berkeley/StABlE-Training}.
\end{abstract}

\section{Introduction}
Molecular dynamics (MD) simulation is a staple method of computational science, enabling high-resolution spatiotemporal modeling of atomistic systems throughout biology, chemistry, and materials science \citep{Frenkel2001MolecularSimulation}.While the atomic forces needed for MD simulation can be obtained on-the-fly via \textit{ab-initio} quantum-mechanical (QM) calculations \citep{Car1985UnifiedApproach}, this is prohibitively expensive for realistic length and time scales \citep{Friesner2005AbInitio}. Machine learning force fields (MLFFs) have recently emerged as a promising option to serve as surrogate models for QM calculations, demonstrating the ability to capture complex many-body interactions, and in some cases transfer flexibly across chemical space \citep{schnet, hu2021forcenet, spherenet, gemnet, Gasteiger2020DirectionalMP, nequip, allegro, Batatia2022MACE, painn}. Graph neural network (GNN)-based MLFFs trained on large \textit{ab-initio} datasets are increasingly being used to model challenging and important chemical systems with favorable results \citep{Merchant2023DeepLearningMaterials, nequipscale, Chen2022UniversalGraphPotential, schaarschmidt2022learned, Takamoto2022UniversalNNPotential, Majewski2023MLProteinThermodynamics, charron2023navigating, mace-mp-0, mace-off23, jmp, gemnetoc, deng2023chgnet}.

\begin{figure}
    \centering
    \includegraphics[width=.9\linewidth]{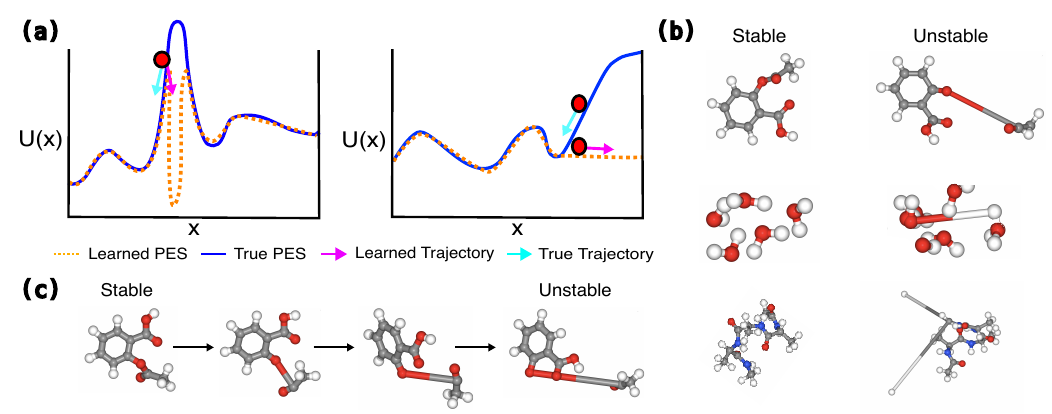}
    \caption{\textbf{Machine learning force field (MLFF) failure modes.} \textbf{(a)} Illustrative examples of true and learned potential energy surfaces (PES) and resulting dynamics for unstable MLFFs. MLFFs can be accurate in approximating \removetwo{most} \addtwo{much} of the PES, but contain \removetwo{localized "holes"} \addtwo{regions} where energy and forces estimates deviate significantly from the true PES, leading to sampling of highly unphysical regimes. As a result, observables computed from MD simulation may be \removetwo{inaccurate} \addtwo{biased by the oversampling of unphysical states,} or have high statistical error \addtwo{in the extreme case of unrecoverable simulation collapse}. \textbf{(b)} Examples of stable versus unstable configurations sampled by MLFFs during molecular dynamics simulation of systems considered in this work. \textbf{(c)} Selected states from an unstable MD trajectory of aspirin.}
    \label{fig:failure_modes}
\end{figure}

MD simulations aim to accurately estimate system observables like the radial distribution function, virial stress tensor, and diffusivity coefficient. This often requires long simulation to fully explore the underlying PES \removetwo{and minimize sampling error}. Unfortunately, MLFFs are known to produce unstable simulations, meaning that they can irreversibly enter unphysical regions of phase space (e.g., a bond breaking event in a non-reactive system at low temperature) \citep{fu2022forces, Stocker_2022, Vita2023DataEfficiency, bihani2023egraffbench, Morrow2023ValidateMLInteratomic, Wang2023MLForceFields}. Sampling of such regions \removetwo{often quickly leads to unrecoverable simulation collapse} \addtwo{can lead to inaccuracies in computed observables} as the MLFF \addtwo{gradually} drifts \removetwo{far} from the distribution of its training data. \addtwo{In extreme cases, instabilities can lead to unrecoverable simulation collapse, in which case} \removetwo{As a result, } computed observables may have high statistical error due to insufficient sampling\removetwo{, or be altogether inaccurate}. \removetwo{This} \addtwo{In either case, instability} can limit the ability of MLFF-based MD simulations to investigate long-timescale phenomena like ion diffusion and protein folding, as well as rare events that may require extensive sampling to encounter. Figure \ref{fig:failure_modes} illustrates typical MLFF instability behaviors and provides examples of unstable structures sampled during MD simulation.

\remove{Insufficient coverage of phase space and inaccuracies within the reference datasets utilized for training MLFFs have been suggested as factors contributing to instability} \add{Simulation instability has been shown to have an unreliable correlation with the energy and force error metrics typically used to train and evaluate MLFFs \citep{fu2022forces, Stocker_2022, bihani2023egraffbench}. Recent works have introduced alternative simulation-based objectives, such as localized \citep{Wang2023MLForceFields} or reweighted \citep{Ge2024SimulationOrientedTraining} energy and force errors, to achieve better downstream stability. However, these approaches do not allow the MLFF to visit new configurations, limiting the distribution over which it is trained and thus the potential improvement. Alternatively, expanding the phase space coverage of the dataset can be an effective way to combat MLFF instability }\citep{fu2022forces, Stocker_2022, Vita2023DataEfficiency, bihani2023egraffbench, Morrow2023ValidateMLInteratomic, Wang2023MLForceFields}. \remove{Expanding the dataset requires additional QM calculations, and improving its accuracy requires more exact QM approaches \citep{Kummel2002CoupledCluster}. Both options are very computationally expensive, particularly for larger systems like macromolecules.} \add{This is typically accomplished via active learning \citep{Smith2018SamplingChemicalSpace, Vandermause2020ActiveLearningForceFields, Schran2020CommitteeNN, Lin2021ActiveLearningPotentials, Kulichenko2023UncertaintyDriven} approaches, where new atomistic configurations are selected, \textit{ab-initio} QM calculations are performed, and the MLFF is retrained on the expanded dataset. However, these techniques rely on performing additional \textit{ab-initio} calculations to expand the dataset. With MLFFs being trained on increasingly diverse datasets \citep{jmp, mace-mp-0, mace-off23} and larger atomistic systems, the expense of these calculations may hamper the practicality of active learning workflows.} This suggests the need for additional sources of information beyond energies and forces to train stable MLFFs. 

In this work, we bridge this gap by using both system observables and \textit{ab-initio} QM data to improve MLFF stability. We introduce \textbf{Stability-Aware Boltzmann Estimator (StABlE) Training}, \remove
two{a procedure designed to produce MLFFs that are both stable and accurate. The core idea behind StABlE is to} \addtwo{which} uses \addtwo{efficient, parallelized} MD simulations to rapidly explore regions of molecular phase space where the MLFF becomes unstable, followed by a targeted refinement of these regions using reference system observables. The key to efficient and numerically stable training lies in the Boltzmann Estimator, which enables end-to-end gradient-based learning without backpropagating directly through MD simulations. We also introduce a localized version of the Boltzmann Estimator, which enables targeted refinement of local instabilities. This is particularly important for stabilizing simulations of large, condensed-phase systems. StABlE Training is a self-contained, efficient process that leverages learning signals from both reference observables and existing QM data, with no reliance on performing additional QM reference calculations. 

We demonstrate StABlE Training on three systems and MLFF architectures: 1) simulation of aspirin with SchNet \citep{schnet}, 2) simulation of the Ac-Ala3-NHMe tetrapeptide with NequIP \citep{nequip}, and 3) simulation of an all-atom water system with GemNet-T \citep{gemnet}. Relative to MLFFs trained solely on energies and forces, our StABlE-trained models produce significantly more stable MD simulations, recover observables more accurately, exhibit better generalization to unseen simulation temperatures, and outperform models trained on 50 times larger, labeled datasets. Our results suggest that utilizing both quantum-mechanical and observable-based modalities is required to fully exploit the available learning signal in reference datasets and train stable and accurate MLFFs. To our knowledge, StABlE Training is the first method that combines these data modalities to improve the stability of neural network potentials in MD simulations.

\section{Preliminaries}
\label{sec:background}

\paragraph{Molecular Dynamics.} 
Molecular dynamics simulation is used to evolve the positions and momenta of an atomistic system. Given a system of $N$ atoms, its state at time $t$ is defined by $\Gamma(t) = \{r(t), p(t)\}$, where $r(t), p(t) \in \mathbb{R}^{N \times 3}$ are the position and momenta of the atoms. We assume the system has a scalar-valued Hamiltonian function $\mathcal{H}: \mathbb{R}^{N \times 3} \times \mathbb{R}^{N \times 3}  \rightarrow \mathbb{R}$ of the form
$ \mathcal{H}(\Gamma) = \sum_{i=1}^{N} \frac{p^{{(i)}^2}}{2m^{(i)}} +  U(r)\,$
where $U: \mathbb{R}^{N \times 3} \rightarrow  \mathbb{R}$ is a potential energy function and $m^{(i)}$ and $p^{(i)}$ are the mass and momentum of atom $i$. By updating in the direction of the per-atom forces $- \frac{\partial U}{\partial r}$ using a numerical integration scheme such as Langevin dynamics \citep{bussi2007accurate}, a sequence of $K$ simulation states $\{\Gamma(t) \}_{t=0}^K$ is produced.

\paragraph{Machine Learning Force Fields.}
A MLFF is a function approximator which learns a potential energy $U_\theta$ and forces $F_\theta = - \nabla_r U_\theta = ( F_{\theta}^{(1)}, \dots, F_{\theta}^{(n)})$ from QM reference data. MLFFs are trained to minimize the following regression loss, with supervision from a dataset of reference energies and forces $\mathcal{D}_{train} = \{(r_i, U_i, F_i)\}_{i=1}^{N}$.

\begin{equation}
\label{eq:bottomup}
\begin{aligned}
\mathcal{L}_{QM} = \frac{1}{N} \sum_{i=1}^N \left[ \lambda_U | U_i(\Gamma) - U_\theta(\Gamma) |^2 + \lambda_F  \sum_{j=1}^{n} \| F_i^{(j)}(\Gamma) + \nabla_{r^{(j)}} U_\theta(\Gamma) \|_2^2 \ \right]
\end{aligned}
\end{equation}

\paragraph{System Observables.} Observables, $g(\Gamma(t))$, characterize the state of a MD simulation at time $t$, and relate to macroscopic properties or experimental measurements of the system. Examples include the radial and angular distribution functions, velocity autocorrelation function, and diffusivity coefficient. Observables can be computed conveniently as an empirical average over states from a MD simulation. This is justified by the ergodicity hypothesis, under which a time-average over a sufficiently long simulation converges to a distributional average over the Boltzmann distribution. More details on the observables used in this work can be found in Supplementary Section \ref{sec: observables}.

\paragraph{Training MLFFs with Observables.} Observables have been used extensively in the historical development of classical potentials \citep{Cornell1995SecondGenForceField, Marrink2007MARTINIForceField, PhysRevB.98.094104}, and more recently are gaining traction as a complementary data source for training MLFFs \citep{wang2020differentiable, difftre, fuchs2025chemtrain}. Observables have been used to train MLFFs in the context of condensed-phase \citep{Wang_2023} and titanium \citep{röcken2023accurate} systems, enhanced sampling of rare events \citep{sipka2023differentiable}, and protein folding simulations \citep{Ingraham2019LearningPS, navarro2023topdown, Kolloff2023dynAMMo}. Training with observables requires an efficient way to compute gradients through MD simulations while avoiding numerical instability and memory limitations. The approach is appealing due to the lack of reliance on expensive \textit{ab-initio} quantum mechanical calculations, and the possibility of improved empirical consistency in settings where the underlying \textit{ab-initio} method may be unreliable \citep{cheetham2024artificial}. However, past works do not consider MLFFs with stability problems, and dispense entirely with using QM data, thus losing the first-principles guarantees of matching an \textit{ab-initio} PES.

 \begin{figure}
    \centering
    \includegraphics[width=0.85\linewidth]{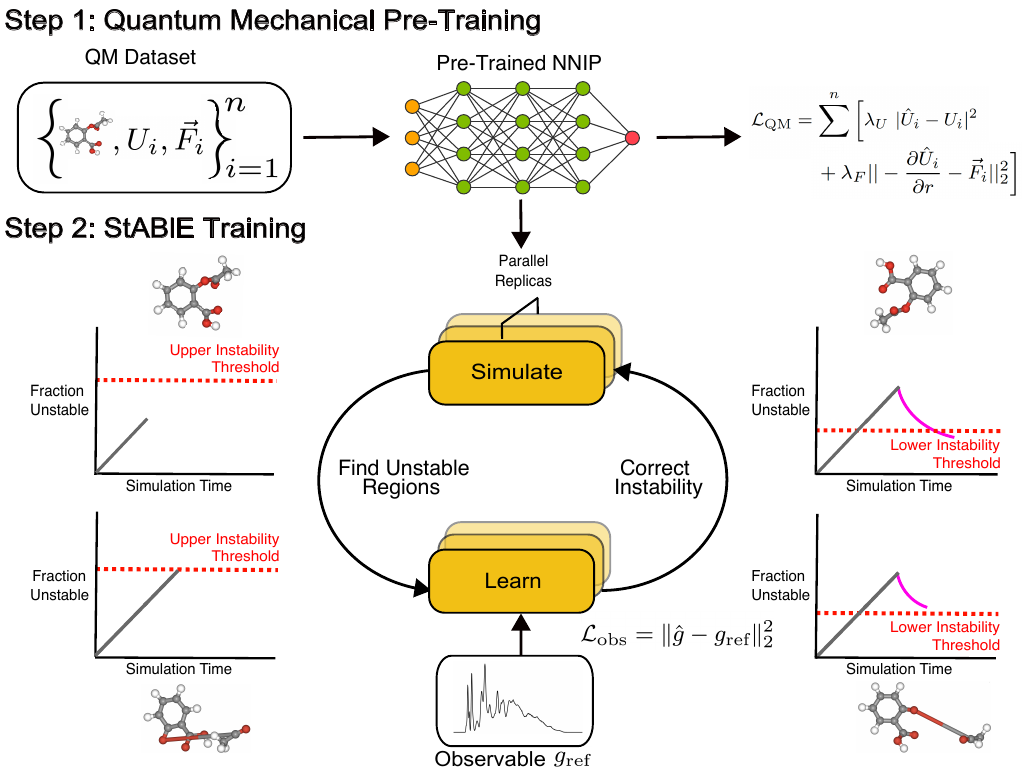}
    \caption{\textbf{Schematic of Stability-Aware Boltzmann Estimator (StABlE) Training procedure.} Our proposed StABlE Training procedure begins with conventional pre-training on a small reference dataset of QM calculations. This dataset remains fixed throughout the procedure, and is never expanded with new calculations. Upon convergence of pre-training, StABlE alternates between two main phases, simulation and learning. In the simulation phase, we perform many molecular dynamics simulations in parallel with the MLFF and find regions of instability. When a sufficient fraction of simulations become unstable, we enter the learning phase, where the MLFF is further trained to match known system observables from \textit{ab-initio} simulation or experiment. Gradients are computed efficiently through the MD simulation via our Boltzmann Estimator. After a sufficient reduction in the portion of unstable trajectories, we re-enter the simulation phase, and repeat the training cycle until a predetermined computational budget is reached.} 
    \label{fig:conceptual}
\end{figure}

\section{Methods}
\label{sec: methods}
We present StABlE Training, our proposed procedure to train stable and accurate MLFFs by leveraging both reference quantum mechanical data and system observables. Our approach is enabled by Boltzmann Estimators, which allows gradient-based optimization of MLFFs based on system observables.
\subsection{Boltzmann Estimator}
\label{sec:boltzmann_estimator}
Reference system observables can be estimated from high-fidelity MD simulations or experimental measurements. Since observables are linked to the MLFF parameters through a molecular simulation, training with this source of information requires a reliable way to optimize through MD trajectories.

Formally, we define $g(\Gamma)$ to be a vector-valued observable of a state $\Gamma$, and $g_{\text{ref}}$ to be the reference value of the observable. To train our MLFF to match a reference observable, we minimize the following loss function:
\begin{equation} \label{eq:topdown}
    \mathcal{L}_{\text{obs}}(\theta) \defas \| \EE_{\Gamma \sim P_\theta(\Gamma)} [g(\Gamma)] - g_{\text{ref}} \|_2^2 ,
\end{equation}
where $P_\theta(\Gamma)$ is the equilibrium distribution induced by the MLFF $U_\theta$. In this work, we primarily consider systems with a fixed volume, temperature, and number of particles, corresponding to the canonical (NVT) ensemble. \add{We note, however, that the estimator is readily applicable to other ensembles, including the isothermal/isobaric (NPT) and grand canonical ($\mu V T$) ensembles (see Supplementary Sections \ref{sec: other_ensembles} \addtwo{and \ref{sec:npt}} for details).}  In the canonical ensemble, microstates obey a Boltzmann distribution, $P_\theta(\Gamma) = {\exp \left(- \frac{1}{k_BT} \mathcal{H}_\theta(\Gamma)\right)} / {C(\theta)}$, where $\mathcal{H}_\theta(\Gamma)$ is the Hamiltonian, $T$ is the sampling temperature, $k_B$ is Boltzmann's constant, and $C(\theta) = \int \exp(- \frac{1}{k_BT}\mathcal{H}_\theta(\Gamma '))\dif \Gamma '$ is the normalizing partition function. MD simulation is required at each training iteration to sample from this distribution and estimate the loss. 

Optimizing the MLFF requires computing gradients of Equation \ref{eq:topdown} with respect to $\theta$. The required gradient, $\nabla_{\theta} \mathcal{L}_{\text{obs}}$, can be decomposed via the chain rule as follows:
\begin{equation}
\begin{aligned} 
    \nabla_{\theta} \mathcal{L}_{\text{obs}}^{\top} &=  \frac{\partial \mathcal{L}_{\text{obs}} } {\partial {\EE_{\Gamma \sim P_\theta(\Gamma)[g(\Gamma)]}}} \frac{\partial \EE_{\Gamma \sim P_\theta(\Gamma)}[g(\Gamma)]}{\partial \theta}
    &= 2 (\EE_{\Gamma \sim P_\theta(\Gamma)} [g(\Gamma)] - g_{\text{ref}} )^{\top} \frac{\partial \EE_{\Gamma \sim P_\theta(\Gamma)}[g(\Gamma)]}{\partial \theta}\,.
    \label{eq:chain_rule}
\end{aligned}
\end{equation}
The non-trivial quantity to compute is the Jacobian, ${\partial \EE_{\Gamma \sim P_\theta(\Gamma)}[g(\Gamma)]}/{\partial \theta}$. 
One way to estimate it is by using a chain rule expansion that corresponds to each step of the unrolled MD simulation \citep{wang2020differentiable}. The adjoint method \citep{chen2019neural} can be used to limit the memory footprint at the expense of increased computation, but can still lead to numerical instability for long trajectories \citep{Wang_2023, sipka2023differentiable} \addblue{(see Supplementary Section \ref{sec:unrolled_comp} for an empirical comparison with direct and adjoint-based backpropagation)}.

We can avoid direct backpropagation through the simulation by noting that the equilibrium state distribution $P_\theta(\Gamma)$ is independent of the algorithm (i.e., MD integrator) used to sample the distribution. This is analogous to implicit differentiation techniques for differentiable optimization \citep{Amos2017OptNet, Gould2016DiffArgminArgmax, ren2022torchopt, Blondel2022EfficientImplicitDiff, negiar2023learning}, in which the solution to an optimization problem is decoupled from the numerical solver used to obtain it. We leverage the known Boltzmann form of $P_\theta(\Gamma)$ to construct an unbiased estimator of the Jacobian.

\begin{center}
    \noindent\fbox{
    \parbox{\textwidth}{
\begin{definition}[$N$-sample Boltzmann estimator]
Given $N$ independent samples $\Gamma_1, \ldots, \Gamma_N$ from a Boltzmann distribution $ P_{\theta}(\Gamma)$, we define the \emph{$N$-sample Boltzmann estimator} $\mathcal{E}(\Gamma_1, \ldots, \Gamma_N)$ \add{of the Jacobian ${\partial \EE_{\Gamma \sim P_\theta(\Gamma)}[g(\Gamma)]}/{\partial \theta}$} as, 
\begin{equation} 
\begin{aligned}
\label{eq:estimator}
&\mathcal{E}(\Gamma_1, \ldots, \Gamma_N) = 
    &\frac{N}{k_BT(N - 1)} \left[\hat{\EE}\left[g(\Gamma)\right]\hat{\EE}\big[{\nabla_\theta U_\theta(\Gamma)}\right]^{\top}\!  - \hat{\EE}\left[{g(\Gamma) \cdot \nabla_\theta U_\theta(\Gamma)}^{\top}\right] \big]\,,
\end{aligned}
\end{equation}
where $\hat{\EE}[f(\Gamma)] = \frac{1}{N}\sum_{i=1}^N f(\Gamma_i)$ denotes an empirical mean over the samples.
\end{definition}

}}
\end{center}

This estimator provides an unbiased estimate of the Jacobian $\frac{\partial \EE_{\Gamma \sim P_\theta(\Gamma)}[g(\Gamma)]}{\partial \theta}$. A proof is provided in Supplementary Section \ref{sec:proofs}. The estimator is related to the REINFORCE trick \citep{williams1992simple} and policy gradient estimators from reinforcement learning \citep{silver2014deterministic} when a Boltzmann state distribution is assumed. The result can also be derived using thermodynamic perturbation theory \citep{Zwanzig1954HighTemperatureEOS, difftre}.

\paragraph{Localized Boltzmann Estimator for Spatial Specificity.} In some scenarios, unphysical configurations can occur within localized regions of the simulation domain, such as collisions between two molecules in a large condensed-phase system. Due to spatial averaging, global observables $g(\Gamma)$ may be insensitive to these localized events, limiting the ability to identify unphysical states.
To address this, we propose the $N$-sample Localized Boltzmann Estimator. Here, the global energy $U_\theta(\Gamma)$ and observable $g(\Gamma)$ are replaced with local versions $U_\theta(\gamma)$ and $g(\gamma)$, where $\gamma$ denotes a local neighborhood of $n < N$ atoms. Formally, define a neighborhood of $n$ atoms $\mathcal{N} = \{ x_1, x_2, \ldots, x_n \mid x_i \in \mathbb{Z}, 1 \leq x_i \leq N \ \text{for all} \ i = 1, 2, \ldots, n \}$. The local state $\gamma$ is defined as $\gamma = \{[r^{(\mathcal{N}_1)}; \ldots ; r^{(\mathcal{N}_n)}], [p^{(\mathcal{N}_1)}; \ldots ; p^{(\mathcal{N}_n})] \}$. The local energy is then defined as $U_\theta(\gamma) = \sum_{i=1}^n U_\theta(\gamma^{(i)})$, where $\gamma^{(i)} = \{r^{(\mathcal{N}_i)}, p^{(\mathcal{N}_i})\}$ contains the position and momenta of the $i^{th}$ atom in the local neighborhood. The local energy is easily obtained by noting that MLFFs parameterize their global energy prediction $U_\theta$ as a sum over individual atomic energies. The localized estimator is thus given as follows:

\begin{center}
    \noindent\fbox{
    \parbox{\textwidth}{
\begin{definition}[$N$-sample localized Boltzmann estimator]

Given $N$ i.i.d. samples of local states $\gamma_1, \ldots, \gamma_N$, where each $\gamma_i$ is extracted from a global state $\Gamma_i \sim P_{\theta}(\Gamma)$, we define the $N$-sample localized estimator of the Jacobian $\frac{\partial \EE_{\gamma \sim P_\theta(\gamma)}[g(\gamma)]}{\partial \theta}$ as

\begin{equation} \label{eq:local_estimator}
    \mathcal{E}(\gamma_1, \ldots, \gamma_N) \defas \frac{N}{k_BT(N - 1)} \left[\hat{\EE}\left[g(\gamma)\right]\hat{\EE}\left[{\nabla_\theta U_\theta(\gamma)}\right]^{\top} - \hat{\EE}\left[{g(\gamma) \cdot \nabla_\theta U_\theta(\gamma)}^{\top}\right] \right]
\end{equation}   
\end{definition}

}}
\end{center}

The localized estimator follows from the original Boltzmann estimator due to the fact that as a subset of the larger Boltzmann-distributed system, any local neighborhood also obeys a Boltzmann-distribution. In practice, we extract multiple local neighborhoods from each global state to increase the sample size and state space coverage (Section \ref{sec: training_details}). A concrete example of localized instability, along with the use of our localized Boltzmann estimator to correct it, will be presented in the context of an all-atom water system in Results, Section \ref{sec:water}.

\paragraph{Key Advantages.}
Unlike active learning approaches, no additional quantum mechanical energy and forces calculations are required to compute our Boltzmann Estimators (all energy terms are model predictions). As a result, learning with the Boltzmann estimator is computationally efficient. Due to the use of independent samples, the estimator also avoids numerical instability and memory demands associated with differentiating through long, continuous trajectories. Further, the most computationally expensive component of the estimator, the gradient of the potential energy $\nabla_\theta U_\theta(\Gamma)$, is independent of the observable $g(\Gamma)$. This means it can be reused, allowing efficient training to match multiple observables simultaneously. 

\subsection{Stability-Aware Boltzmann Estimator (StABlE) Training}
\label{sec:stable}
Stability-Aware Boltzmann Estimator (StABlE) Training begins with conventional supervised pre-training of a MLFF on a reference dataset of energy and forces, using the loss defined in Equation \ref{eq:bottomup}. The method then proceeds by alternating between two major phases, simulation and learning, as illustrated in Figure \ref{fig:conceptual}.

\paragraph{Simulation Phase.}
The simulation phase aims to explore the molecular phase space and pinpoint regions where the MLFF becomes unstable.
We sample $R$ equilibrium states from the training dataset as initial conditions for separate MD trajectories (replicas). 
Using the pre-trained MLFF $U_\theta$, we run MD simulations on these replicas in parallel for $t$ timesteps. \add{Due to our use of efficient vectorized GPU operations, simulating many replicas in parallel has similar computational cost to simulating a single replica.}
We apply a predetermined stability criterion (Supplementary Section \ref{sec: stability_criteria}) to each replica, freezing those marked unstable at their current states.
Simulations continue for the remaining replicas, with stability checks every $t$ timesteps. Once a specified fraction of replicas become unstable, we rewind all unstable replicas by $t$ timesteps. This ensures that further simulation for $t$ timesteps will trigger instability, which is then corrected in the next phase. \add{We note that many other strategies exist to explore phase space and find unstable regions, including Diffusion Monte Carlo with fictitious masses \citep{Li2021DiffusionMonteCarlo} or uncertainty-based sampling \citep{Kulichenko2023UncertaintyDriven}. StABlE Training can be flexibly used with any of these strategies without impacting the learning stage (described next) of the procedure.}

\paragraph{Learning Phase.}
The learning phase aims to refine the MLFF to correct the previously encountered instabilities. Starting from the near-unstable configurations obtained in the simulation phase, we perform MD simulation for $t$ timesteps, sampling every $S^{th}$ state to obtain uncorrelated samples. Using the sampled states $\Gamma_1,\ldots, \Gamma_{N_d}$, where \remove{$N_d = \frac{t}{S}$} \add{$N_d = \frac{tR}{S}$}, we compute the observable loss function (Equation \ref{eq:topdown}). To update the MLFF parameters $\theta$, we compute an unbiased estimator of the loss gradient via the Boltzmann Estimator (Section \ref{sec:boltzmann_estimator}) and use it to perform a single step of gradient descent. We then reset all replicas to their original near-unstable states, simulate with the updated MLFF, recompute the loss and gradient estimator with the newly sampled states, and again update the MLFF weights. This process is repeated until the fraction of unstable replicas drops below a predetermined threshold. When this occurs, the learning phase has concluded, and we begin a new simulation phase starting from the endpoints of the last learning phase. We continue alternating between simulation and learning phases until a predetermined computational budget is reached, at which point StABlE Training has concluded. See Supplementary Section \ref{sec: algorithm} for a formal algorithmic description of the StABlE Training procedure.

\paragraph{Regularizing StABlE Training with Energy and Forces Reference Data.}
In practice, the mapping between a sparse set of system observables and a potential energy function is non-unique \citep{Noid2013CoarseGrainedModels}. Consequently, learning with observables alone is underconstrained. To combat this, we regularize the observable loss function (Equation \ref{eq:topdown}) with the energy and forces loss function (Equation \ref{eq:bottomup}). The final StABlE loss function thus becomes,
\begin{equation}
    \mathcal{L}_{\text{StABlE}}(\theta) \defas \mathcal{L}_{\text{obs}} + \lambda \mathcal{L}_{\text{QM}}, 
\end{equation}
where $\mathcal{L}_{\text{obs}}$ and $\mathcal{L}_{\text{QM}}$ were defined in Equations \ref{eq:topdown} and \ref{eq:bottomup} respectively, and $\lambda$ controls the strength of the regularization. Crucially, $\mathcal{L}_{\text{QM}}$ is only computed over the original training dataset $\mathcal{D}_{train}$, and not over new structures explored during MD simulation. Therefore, the regularization requires no additional \textit{ab-initio} calculations.
\section{Results}
\label{sec:results}

We present the results of StABlE Training on the aspirin molecule with SchNet (Section \ref{sec:aspirin}), Ac-Ala3-NHMe tetrapeptide with NequIP (Section \ref{sec:ac_ala3_nhme}), and an all-atom water system with GemNet-T (Section \ref{sec:water}).

\subsection{Aspirin Molecule}
\label{sec:aspirin}

\begin{figure}
    \centering
    \includegraphics[width=0.85\linewidth]{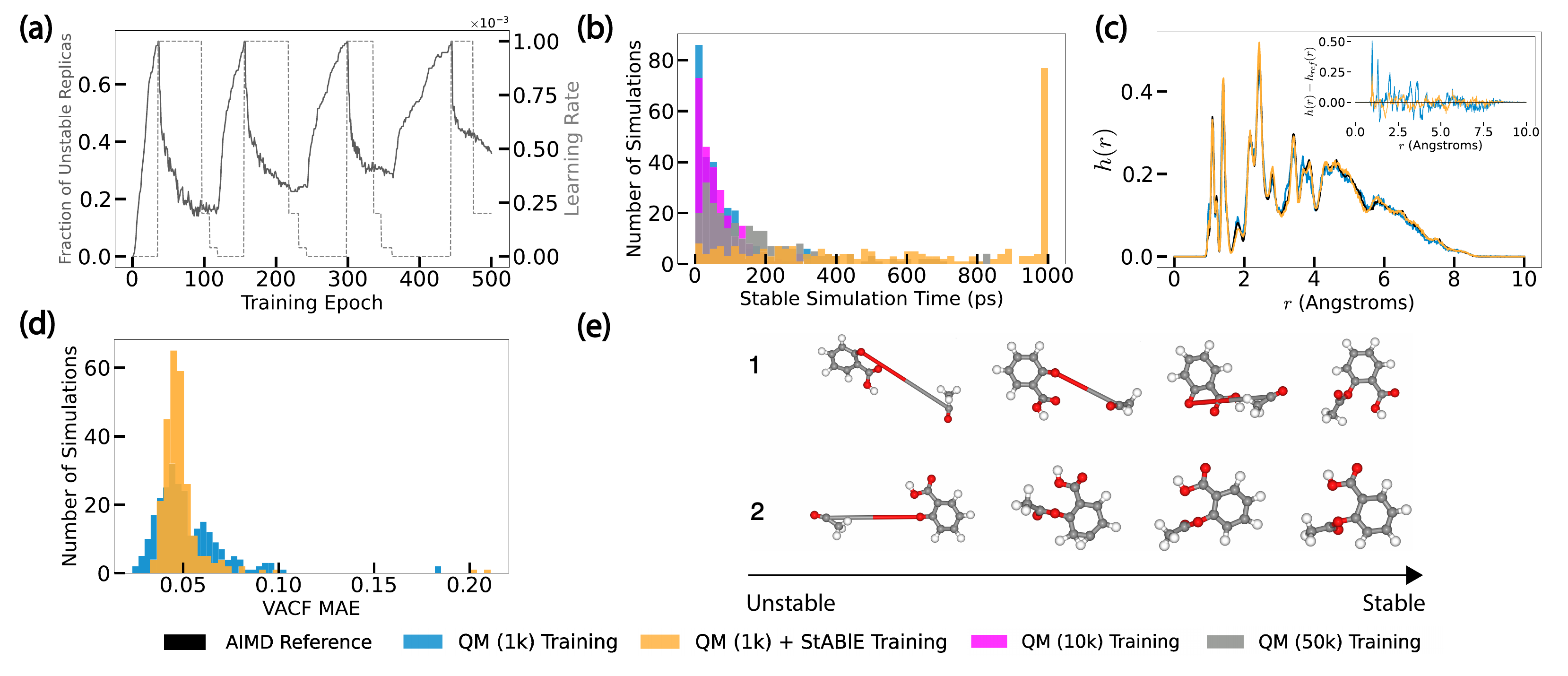}
    \caption{\textbf{Aspirin simulation with StABlE Training.} \textbf{(a)} Alternation  of simulation and learning phases during StABlE Training with 128 parallel replicas. The simulation phases correspond to regions where the fraction of unstable replicas increases, while the learning phases correspond to regions where learning occurs and the fraction of unstable replicas decreases. \textbf{(b)} Stable simulation time of 256 parallel aspirin trajectories from SchNet MLFFs. Applying StABlE Training yields significantly more stable simulations than models trained only on energies and forces, surpassing models trained on 50$\times$ more QM data. \textbf{(c)} Distribution of interatomic distances ($h(r)$) from MLFF simulations. A StABlE-trained SchNet model closely recovers the true distribution of interatomic distances, while the model trained only on QM reference data produces a noisier $h(r)$ because it cannot stably simulate the system for longer time periods. Inset shows difference between predicted and reference $h(r)$. \textbf{(d)} Distribution of velocity autocorrelation function (VACF) mean absolute error (MAE). StABlE Training yields a reduction in variance across replicas. \textbf{(e)} Aspirin structures sampled over epochs of a single learning phase of StABlE Training. There is a clear progression as unstable configurations become stable.}
    \label{fig:md17_result}
\end{figure}

Aspirin (chemical formula $C_9H_8O_4$) is the largest molecule from MD17 \citep{md17}, a widely used benchmark dataset for atomistic simulations which contains energy and forces calculations computed at the PBE+vdW-TS \citep{md17_dft} level of theory. Consisting of 21 atoms, aspirin has been shown to be the most challenging molecule in MD17 for state-of-the-art MLFFs to simulate stably~\citep{fu2022forces}. We pre-train a SchNet \citep{schnet} model on the energy and forces matching objective (Equation \ref{eq:bottomup}) using a subset of 1,000 aspirin structures from the reference dataset. After convergence, we begin StABlE Training with the global Boltzmann estimator, using 128 parallel replicas at a temperature of 500K. By simulating hundreds of replicas in parallel, we expose the MLFF to a comprehensive range of failure modes much more quickly than possible with a single replica. Following \citep{fu2022forces}, we use a maximum bond length deviation criterion to detect instability in the simulations (Section \ref{sec: stability_criteria}). We use the distribution of interatomic distances, $h(r)$, as our training observable, with the $h(r)$ computed over structures in the training dataset serving as the reference (Section \ref{sec: observables}). This choice is motivated by the observation that unphysical bond stretches constitute the majority of failure cases in aspirin simulations. The $h(r)$ is sensitive to these abnormalities, and is thus an informative optimization criterion. We perform four complete StABlE cycles of simulation and learning with a SGD optimizer. We evaluate our final models by performing \addtwo{1 nanosecond} constant-temperature MD simulations, starting from 256 initial structures not seen during training. Figure \ref{fig:md17_result}a shows the progression of StABlE Training. During the simulation phase, the learning rate is zero and the fraction of unstable replicas steadily increases. When the upper instability threshold is reached, learning commences. The fraction of unstable replicas steadily decreases, confirming that optimization of the interatomic distance distribution produces the desired stability improvement.

\begin{figure}
    \centering
    \includegraphics[width=0.72\linewidth] {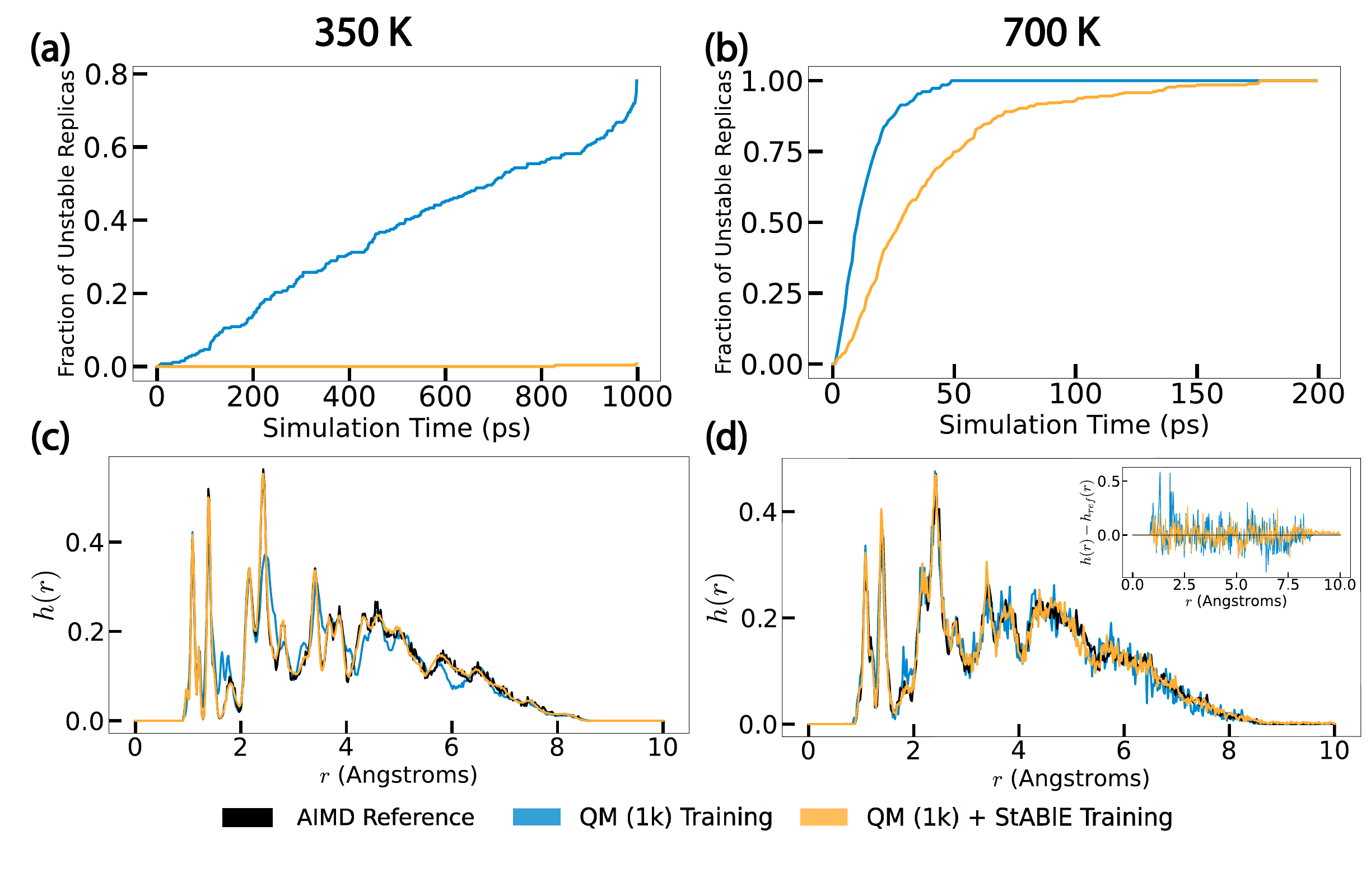}
    \caption{\textbf{Testing temperature generalization for aspirin.} \textbf{(a, b)} Fraction of unstable replicas as a function of simulation time for SchNet MLFFs at 350K and 700K. Applying StABlE Training yields significantly more stable simulations than a baseline SchNet model. \textbf{(c, d)} Distribution of interatomic distances ($h(r)$) from SchNet MLFFs. The StABlE-trained SchNet model more accurately recovers the true $h(r)$ at both 350K and 700K (the inset shows difference between the predicted and reference $h(r)$). The AIMD reference $h(r)$ is computed by Boltzmann-reweighting of independent samples from the 500K training dataset.} 
    \label{fig:md17_temperature_result}
\end{figure}

A SchNet model trained via our stability-aware approach is significantly more stable during MD simulation than models trained only on the conventional QM energy and forces objective function (Figure \ref{fig:md17_result}b). As a result of our training procedure, the \removetwo{mean} \addtwo{median} stable simulation time increases from 42 to 602 picoseconds. Even when StABlE training is employed after training on only on 1,000 reference structures, it is more stable than models trained conventionally on 10,000 and 50,000 reference structures, highlighting the effectiveness of our method in improving stability without reliance on any additional QM reference data. We also observe in Supplementary Section \ref{sec:dt_analysis} that simply reducing the simulation timestep \addtwo{by a factor of ten} does not eliminate instability in simulations produced by conventionally trained models, \addtwo{and reducing instability further may require impractically small timesteps}. \addblue{In Supplementary Section \ref{sec:dt_analysis}, we demonstrate the potential of StABlE Training to accelerate simulations by enabling larger timesteps.} Simulations from our model closely recover the true distribution of interatomic distances, while the distributions produced by models solely trained on QM data are noisier, due to the limited stable simulation time (Figure \ref{fig:md17_result}c). 
We also test the ability of our StABlE-trained model to recover the velocity autocorrelation function (VACF), a fundamental dynamical observable not seen during training (Figure \ref{fig:md17_result}d). We see that the quality of the recovered VACF remains similar after StABlE Training, with lower variance over the simulated replicas (see Supplementary Section \ref{sec:vacf} for a sample VACF computed from the simulations). The preservation of a held-out dynamical observable suggests that StABlE Training does not achieve stability improvements by simply restricting the model around a narrow regime corresponding to the reference $h(r)$. If this were the case, characteristic dynamic fluctuations around the reference $h(r)$ would be suppressed, leading to an inaccurate VACF. Figure \ref{fig:md17_result}e depicts aspirin structures sampled from simulations during the course of a single learning phase of StABlE Training, demonstrating that unphysical bond stretching in these structures is resolved by the training procedure.

\subsection{Temperature generalization}
\label{sec:aspirin_temperature}

We explore our method's generalization to different thermodynamic conditions. To do so, we perform 256 parallel MD simulations at 350K and 700K using the SchNet model which was trained with our StABlE Training procedure at 500K. We use the same criterion based on bond length deviation to detect instability as in Section \ref{sec:aspirin}. We estimate the reference $h(r)$ at 350K and 700K by Boltzmann-reweighting samples from the original 500K training dataset (Section \ref{sec: temperature_reweighting}). Aspirin simulations produced by a StABlE-Trained SchNet model are significantly more stable at both temperatures than baseline models trained only on QM reference data (Figure \ref{fig:md17_temperature_result}a, b). At 350K, the difference is particularly large. After 1 nanosecond of simulation, \removetwo{less than 5\% of} \addtwo{virtually no} replicas are unstable in the StABlE-trained MLFF simulation, while approximately 80\% of replicas are unstable in the baseline MLFF simulation. The StABlE-trained model also recovers the true distribution of interatomic distances more closely than the baseline MLFF at both temperatures (Figure \ref{fig:md17_temperature_result}c, d). At 350K, the baseline MLFF is stable enough produce a smooth $h(r)$, but the structure is inaccurate, indicating incorrect sampling of the phase space. At 700K, the baseline MLFF produces a $h(r)$ which is close to the AIMD-produced distribution, but is noisy due to insufficient stable simulation time. This underscores the necessity for a MLFF to be both stable and accurate in order to be practically useful in MD simulation. StABlE Training improves both of these metrics for this system. As generalization to new thermodynamic conditions is of crucial importance for MLFFs to be useful in practical applications \citep{3bpa}, we see the temperature-transferability of StABlE Training as an important strength of our method.

\subsection{Ac-Ala3-NHMe Tetrapeptide}
\label{sec:ac_ala3_nhme}

\begin{figure}
    \centering
\includegraphics[width=0.85\linewidth] {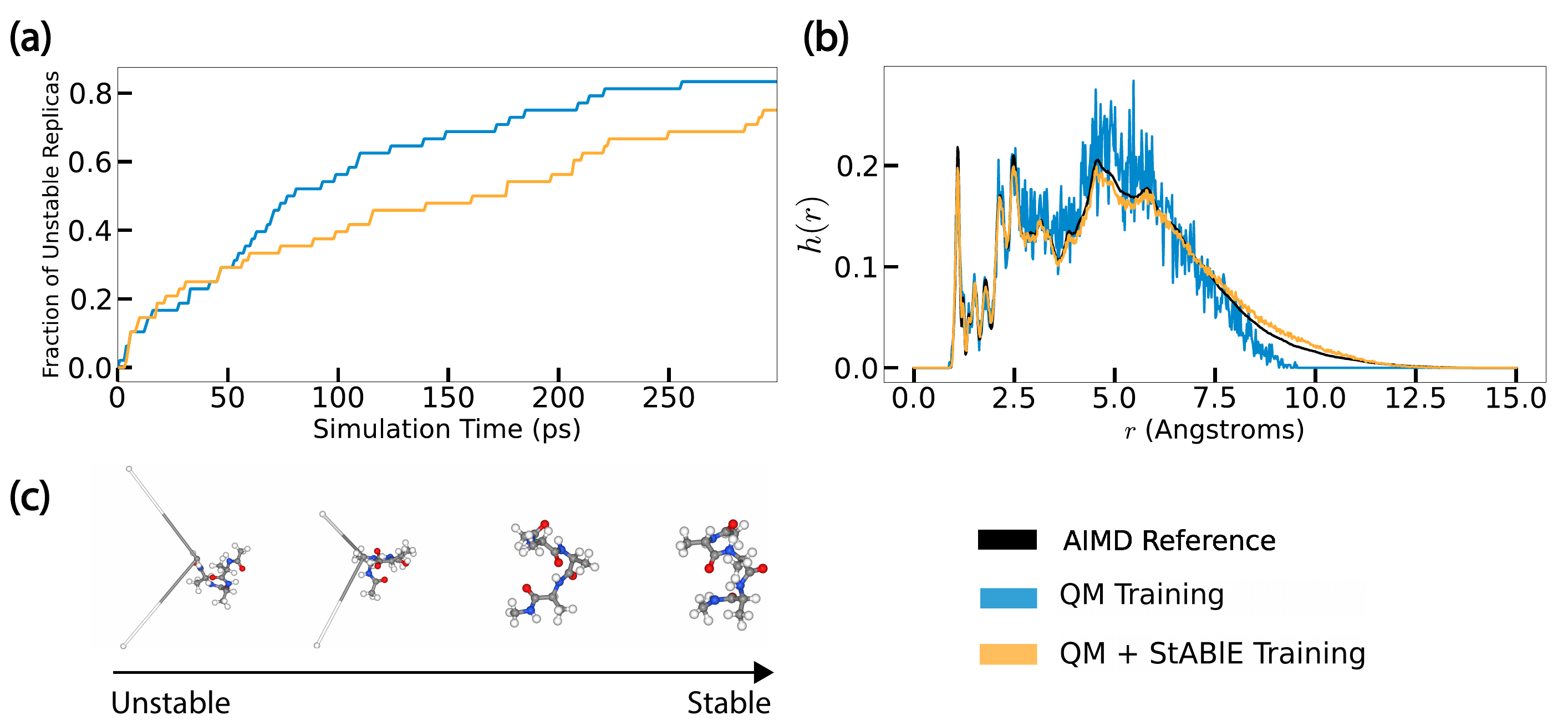}
    \caption{\textbf{Ac-Ala3-NHMe tetrapeptide simulation with StABlE Training.}\textbf{(a)} Fraction of unstable replicas as a function of simulation time for NequIP MLFFs. Applying StABlE Training yields a model which can simulate more MD replicas stably over time than a baseline trained only on energies and forces. 
\textbf{(b)} Distribution of interatomic distances ($h(r)$) from NequIP MLFF simulations. The StABlE-trained NequIP model much more closely recovers the true $h(r)$, while the $h(r)$ produced by the model trained only on QM reference data is noisy and inaccurate due to insufficient sampling time. \textbf{(c)}  
Ac-Ala3-NHMe structures sampled over epochs of a single learning phase of StABlE Training. There is a clear progression as very unstable configurations become stable.}
    \label{fig:md22_result}
\end{figure}

Ac-Ala3-NHMe (chemical formula $C_{12}H_{22}N_4O_4$) is a tetrapeptide from the MD22 dataset \citep{md22}, a challenging benchmark consisting of reference energy and forces computed at the PBE+MBD \citep{md22_dft1, md22_dft2} level of theory for considerably larger molecules than those in the MD17 dataset. Relative to MD17 molecules, Ac-Ala3-NHMe poses unique challenges for atomistic simulations due to its larger size and flexibility \citep{Kabylda2023EfficientInteratomicDescriptors}. As a result, more expressive MLFFs---sometimes incorporating E(3) equivariance \citep{geiger2022e3nn}---are required to accurately model the underlying potential energy surface. In our experiments, we use a NequIP \citep{nequip} model due to its promising accuracy and data efficiency on challenging atomistic simulation tasks \citep{nequipscale, Merchant2023DeepLearningMaterials}. We pre-train a NequIP model on the QM energy and forces matching objective using 14,890 (or 25\%) of available structures from the training dataset. After convergence, we begin StABlE Training with the global Boltzmann estimator, using 128 parallel replicas at a temperature of 500K. As with aspirin (Sections \ref{sec:aspirin} - \ref{sec:aspirin_temperature}), we detect instability using the maximum bond length deviation criterion, and choose the distribution of interatomic distances, $h(r)$, as our training observable. The StABlE-trained NequIP model gives an improvement in stability relative to a baseline model trained only on QM reference data (Figure \ref{fig:md22_result}a). Additionally, the StABlE-trained model closely recovers the true $h(r)$, while the baseline MLFF produces a highly noisy $h(r)$. This is because of insufficient sampling, due to a lower stable simulation time (Figure \ref{fig:md22_result}b). The baseline MLFF also underestimates the latter portion of the $h(r)$ distribution, indicating poor modelling of long-range interactions. Figure \ref{fig:md22_result}c shows selected structures throughout the course of a single learning phase of StABlE Training, demonstrating a progression from very unstable to stable structures.

\subsection{Water}
\label{sec:water}

\begin{figure}
    \centering
\includegraphics[width=0.9\linewidth] {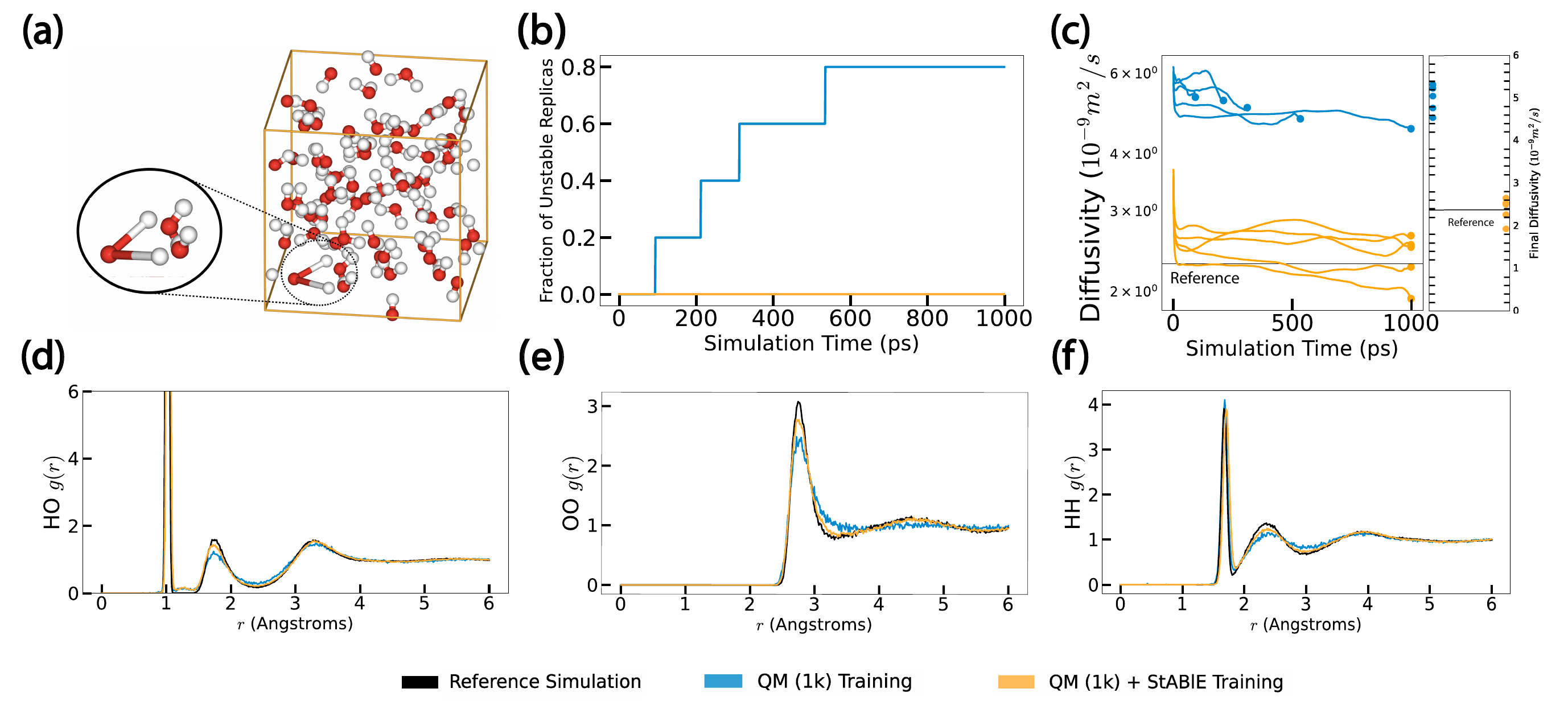}
    \caption{\textbf{All-Atom Water Simulation with StABlE Training.} \textbf{(a)} Example unstable configuration arising in a GemNet-T simulation of water. Instabilities are highly localized and manifest as unphysical bond stretches or intermolecular collisions, motivating the use of our localized Boltzmann estimator (Section \ref{sec:boltzmann_estimator}) during training  \textbf{(b)} Fraction of unstable replicas as a function of simulation time for GemNet-T MLFFs. StABlE Training yields a significant stability improvement relative to training on only energy and forces data. \textbf{(c)} Convergence of the diffusivity coefficient over MLFF simulations. The StABlE-trained model gives a considerably more accurate final estimate of the diffusivity coefficient. \textbf{(d - f)} Element-conditioned (OH, OO, HH) radial distribution functions (RDF) produced by MLFF simulations. The StABlE-trained model \removetwo{produces fewer unphysical collisions and} more accurately captures long-range interactions. 
} 
    \label{fig:water_result}
\end{figure}

As a final evaluation of StABlE Training, we consider liquid-phase water. Water has historically posed unique challenges for atomistic simulations due to the presence of long-range interactions, proton disorder, and nuclear quantum effects \citep{Cheng2019AbInitioWater, Raza2011ProtonOrderingIce, Markland2018NuclearQuantumEffects}. Additionally, estimating transport properties, such as the diffusivity coefficient,  requires long, continuous trajectories to minimize statistical error. We use the dataset produced in Fu et al. \citep{fu2022forces}, which contains reference all-atom simulations of 64 water molecules using the flexible Extended Simple Point Charge model \citep{Wu2006FlexibleSPC} at 300K and 1 atm. For this system, we consider a GemNet-T model \citep{gemnet}. GemNet-T displays a failure mode in which unphysical configurations (e.g., bond stretching) first arise in highly localized "pockets" of 1-2 molecules (Figure \ref{fig:water_result}a), which then gradually cascade to the rest of the simulation domain \citep{fu2022forces}. Such configurations do not noticeably affect global system observables such as the radial distribution function. Therefore, in this setting it is more appropriate to use our localized version of the Boltzmann gradient estimator (Equation \ref{eq:local_estimator}) in order to guide the optimization process towards specific spatial domains of instability. After pre-training a GemNet-T model using a subset of 1,000 reference structures, we begin StABlE Training with the localized Boltzmann estimator, simulating 5 parallel replicas at a temperature of 300K. We use a minimum intermolecular distance criterion (Section \ref{sec: stability_criteria}) to detect instabilities \addblue{during training}. To estimate gradients for optimization, from each global state we extract all possible local neighborhoods containing a single water molecule (Supplementary Section \ref{sec: training_details}). The most prevalent failure mode \addtwo{produced by GemNet-T} in this system is unphysical bond stretching, which eventually leads to unphysical coordination structures. Motivated by this observation, we use the mean hydrogen-oxygen bond length as our training observable. We perform a single cycle of simulation and learning. We then evaluate the performance of the StABlE-Trained model by performing 5 parallel nanosecond (ns) MD simulations at 300K, starting from held-out initial conditions \addblue{and using a RDF MAE criterion (defined in Supplementary Section \ref{sec: stability_criteria}) to measure stability}. We achieve significant improvements in stability and accuracy using our training approach. \removetwo{We nearly double the stable simulation time relative to a baseline MLFF model trained only on QM reference data} \addtwo{Our StABlE-trained model can simulate stably for 1 ns for all 5 initial conditions, while the baseline MLFF model trained only on QM reference data can only do so for 1 initial condition, and has a median stability of just 312 ps} (Figure \ref{fig:water_result}b). Compared to a reference value of $2.3 \times 10^{-9} \frac{m^2}{s}$, the mean diffusivity produced by StABlE-trained model simulations is \addtwo{$2.4 \times 10^{-9} \frac{m^2}{s}$}, while the mean diffusivity produced by baseline model simulations is $5.0 \times 10^{-9} \frac{m^2}{s}$ (Figure \ref{fig:water_result}c). The \addtwo{long-range correlations in the element-conditioned RDFs (Figures \ref{fig:water_result}d-f) are also captured more accurately by the StABlE-trained model}. We reiterate that neither the element-conditioned RDFs nor the diffusivity coefficient were explicitly seen during StABlE Training. \addblue{We also highlight that the RDF MAE stability criterion used for evaluation is different from the minimum intermolecular distance criterion used during training (see Supplementary Section \ref{sec: stability_criteria} for more details), suggesting that StABlE Training does not overfit to the training stability criterion.} \addtwo{In Supplementary Section \ref{sec:npt}, we demonstrate that StABlE Training leads to similar stability improvements when water simulations are conducted in the isothermal/isobaric (NPT) ensemble, instead of the canonical (NVT) ensemble.}
\section{Conclusion and Future Work}
\label{sec:conclusion}

We have introduced StABlE Training, a strategy for training stable and accurate machine learning force fields. Our training procedure results in MLFFs which are significantly more stable in MD simulation, more accurately reproduce key simulation observables, (including those that were not explicitly trained on), exhibit better generalization to unseen temperatures, and have superior data efficiency relative to MLFFs trained only on QM data.   

\paragraph{Key Takeaways.} StABlE Training can utilize both quantum-mechanical energies and forces and system observables to supervise MLFF training. \addtwo{We highlight that the reference observables need not be acquired from experimental measurements, and can be computed by averaging over existing QM datasets. This makes our approach applicable in realistic computational discovery scenarios in which experimental characterization is not always available for hypothetical systems.} \addtwo{The} stability, accuracy, and data efficiency gains brought by StABlE Training require \textit{no additional reference calculations or data}. \addtwo{The procedure is thus} self-contained and efficient, requiring minimal additional computational expense beyond \add{a single iteration of} conventional MLFF training (see Supplementary Section \ref{sec: training_details} for details on training times). \add{This suggests that StABlE Training may scale more gracefully to larger MLFFs and atomistic systems than existing active learning approaches, which require repeated QM reference calculations and retraining.} 
StABlE Training can also be flexibly applied across atomistic systems due to its ability to handle a diverse set of failure modes arising in MD simulation, including both global and local instabilities due to our localized Boltzmann estimator. This flexibility extends to the choice of MLFF: the effectiveness of StABlE Training across the three diverse architectures considered in this work suggests that our approach will remain applicable as MLFF architectures continue to evolve.

\paragraph{Limitations.} Unlike traditional QM learning, training MLFFs to match reference observables lacks convergence guarantees in the large data limit. Specifically, the observable-matching objective is under-constrained, as the mapping between a potential energy function and a sparse set of simulation observables is in general non-unique \citep{Noid2013CoarseGrainedModels}. Although we mitigate this problem by pre-training and regularizing StABlE Training with the conventional energy and forces loss function, we observe that stability improvements resulting from StABlE Training are accompanied by a minor increase in the energy and forces error on a held-out test set, indicating a misalignment between the observable-matching and QM objectives. Adjusting the strength of the QM regularization and the learning rate used for StABlE Training are two primary ways in which to navigate this tradeoff. We note, however, that our observed error increases are typically within the range of DFT error (Supplementary Section \ref{sec:energy_force_analysis}). We also note that StABlE Training is currently incompatible with dynamical observables, due to the use of uncorrelated states to compute the Boltzmann Estimator. Overcoming this limitation is nontrivial as it requires optimization over long paths, but may become tractable with recent advances in path gradient computation \citep{Bolhuis2023OptimizingMolecularPotentials, greener2024reversible, han2025refining}.

\paragraph{Future Work.} In this work, we took observables calculated from high-fidelity simulations as reference values. Many of the observables we considered, such as the radial distribution function, diffusivity coefficient, and equilibrium bond length, are also experimentally measurable. Future work could explore training with experimental observables from multiple thermodynamic conditions simultaneously to yield more generalizable and robust potentials. Future work could also explore incorporating additional observables into StABlE Training, particularly those which are dynamical. This could restrict the set of learnable functions and address the under-constrained nature of learning with observables.  \addtwo{We also note that the reference datasets used to pre-train MLFFs in this work were sampled uniformly from high-fidelity simulations. More sophisticated sampling strategies \citep{deringer2018data, karabin2020entropy, yoo2021metadynamics, fonseca2021improving, Kulichenko2023UncertaintyDriven, qi2024robust} for diverse dataset generation, as well as during phase space exploration to find unstable regions, could be employed along with StABlE Training to achieve further stability and accuracy gains.} Closely related techniques such as active learning are \addtwo{compatible} with StABlE Training, and could be used in tandem to further improve MLFF stability and robustness. \add{In this setting, system observables can serve as a cheap source of information with which to augment the more expensive supervision provided by \textit{ab-initio} calculations.}

    

\subsubsection*{Acknowledgments}
This work was supported by the U.S. Department of Energy, Office of Science, Office of Advanced Scientific Computing Research, Scientific Discovery through Advanced Computing (SciDAC) program under contract No. DE-AC02-05CH11231, and the U.S. Department of Energy, Office of Science, Energy Earthshot initiatives as part of the Center for Ionomer-based Water Electrolysis at Lawrence Berkeley National Laboratory under Award Number DE-AC02-05CH11231. 
This research used resources of the National Energy Research Scientific Computing Center (NERSC), a U.S. Department of Energy Office of Science User Facility located at Lawrence Berkeley National Laboratory, operated under Contract No. DE-AC02-05CH11231. We thank Rasmus Lindrup, Toby Kreiman, Geoffrey Negiar, Muhammad Hasyim, Ritwik Gupta, Martin Sipka, Johannes Dietschreit, Aayush Singh, David Limmer, Kranthi Mandadapu, David Prendergast, Bryan McCloskey, and Muratahan Aykol for fruitful discussions and comments on the manuscript.

\bibliography{main}
\bibliographystyle{tmlr}

\newpage

\appendix
\section{Appendix}

\subsection{Derivation of Boltzmann Estimator}\label{sec:proofs} We provide a full derivation of our Boltzmann estimator, which we use to train our MLFF as part of StABlE Training. Consider a vector-valued observable $g(\Gamma)$ of a state $\Gamma$, and a reference value of the observable $g_{\text{ref}}$. Training a MLFF $U_\theta$ to match $g_{\text{ref}}$ requires minimizing the loss function, 
$$
    \mathcal{L}_{\text{obs}}(\theta) \defas \| \EE_{\Gamma \sim P_\theta(\Gamma)} [g(\Gamma)] - g_{\text{ref}} \|_2^2,
$$
where $P_\theta(\Gamma)$ is the equilibrium distribution induced by the MLFF $U_\theta$. This requires computing the gradient, $\nabla_{\theta} \mathcal{L}_{\text{obs}}$, which can be decomposed via the chain rule as follows:
$$
\begin{aligned} 
    \nabla_{\theta} \mathcal{L}_{\text{obs}}^{\top} &= \frac{\partial \mathcal{L}_{\text{obs}} }{\partial {\EE_{\Gamma \sim P_\theta(\Gamma)[g(\Gamma)]}}} \frac{\partial \EE_{\Gamma \sim P_\theta(\Gamma)}[g(\Gamma)]}{\partial \theta} \\
    &= 2 (\EE_{\Gamma \sim P_\theta(\Gamma)} [g(\Gamma)] - g_{\text{ref}} )^{\top} \frac{\partial \EE_{\Gamma \sim P_\theta(\Gamma)}[g(\Gamma)]}{\partial \theta}\,.
    \label{eq:chain_rule_appendix}
\end{aligned}
$$

We derive the $N$-sample estimator, presented in Equation 6 in the main text, of the Jacobian, ${\partial \EE_{\Gamma \sim P_\theta(\Gamma)}[g(\Gamma)]}/{\partial \theta}$. The estimator is repeated below for convenience.

$$
    \mathcal{E}(\Gamma_1, \ldots, \Gamma_N) \defas \frac{N}{k_BT(N - 1)} \left[\hat{\EE}\left[g(\Gamma)\right]\hat{\EE}\left[{\nabla_\theta U_\theta(\Gamma)}\right]^{\top} - \hat{\EE}\left[{g(\Gamma) \nabla_\theta U_\theta(\Gamma)}^{\top}\right] \right] ,
$$       
where $\hat{\EE}$ denotes sample averages. This is an unbiased estimator, that is, $
\EE_{\Gamma_1, \ldots, \Gamma_N \sim P_{\theta}(\Gamma)} [\mathcal{E}(\Gamma_1, \ldots, \Gamma_N)] = \frac{\partial \EE_{\Gamma \sim P_\theta(\Gamma)}[g(\Gamma)]}{\partial \theta}$.

\begin{proof}
    Given $P_\theta(\Gamma) \defas \frac{\exp \left(- \frac{1}{k_BT} \mathcal{H}_\theta(\Gamma)\right)}{C(\theta)}$, where $\mathcal{H}_\theta(\Gamma) = \sum_{i=1}^{N} \frac{p_i^2}{2m_i} +  U_\theta(r)$, and $C(\theta) \defas \int \exp(- \frac{1}{k_BT}\mathcal{H}_\theta(\Gamma '))\dif \Gamma '$ is the partition function, we have,
    \begin{align*}
    \frac{\partial \EE_{\Gamma \sim P_\theta(\Gamma)}[g(\Gamma)]}{\partial \theta}  & =\frac{\partial}{\partial \theta} \int g(\Gamma') P_\theta(\Gamma')\dif \Gamma '   \\ 
    & =\int g(\Gamma ') \nabla_\theta P_\theta(\Gamma ')^{\top} \dif \Gamma ' \\
    &= \int g(\Gamma ') \nabla_\theta (\frac{\exp \left(- \frac{1}{k_BT} \mathcal{H}_\theta(\Gamma)\right)}{C(\theta)})^{\top} \dif \Gamma '. \\
    \end{align*}
    Expanding via the chain rule for Jacobians, and noting that $ \nabla_{\theta} \mathcal{H}_{\theta}(\Gamma) = \nabla_{\theta} U_{\theta}(r)$ (we will write $U_{\theta}(\Gamma)$ for convenience) \add{since the kinetic energy is independent of $\theta$}, we get,
    
    \begin{align*}
    \add{\frac{\partial \EE_{\Gamma \sim P_\theta(\Gamma)}[g(\Gamma)]}{\partial \theta}} & = \int g(\Gamma ') \frac{- \frac{1}{k_BT}\exp \left(- \frac{1}{k_BT} \mathcal{H}_{\theta}(\Gamma ')\right) \nabla_{\theta} U_{\theta}(\Gamma ')^{\top} C(\theta) - \nabla_\theta C(\theta)^{\top} \exp \left(- \frac{1}{k_BT} \mathcal{H}_{\theta}(\Gamma ')\right)}{C(\theta)^2} \mathrm{d} \Gamma ' \\
    & =\int g(\Gamma ')\left[- \frac{1}{k_BT}  \nabla_\theta U_\theta(\Gamma ')^{\top} P_\theta(\Gamma ')-  \frac{\nabla_\theta C(\theta)^{\top}}{C(\theta)}P_\theta(\Gamma ')\right] \mathrm{d} \Gamma ' \,.
    \end{align*}
    
    By definition of $C(\theta)$, the quotient $\frac{\nabla_\theta C(\theta)}{C(\theta)}$ can be simplified as,
    
    $$
    \begin{aligned}
    \frac{\nabla_\theta C(\theta)}{C(\theta)} & =\frac{- \int \nabla_\theta U_\theta(\Gamma ') \exp \left(- \frac{1}{k_BT} \mathcal{H}_{\theta}(\Gamma ')\right) \mathrm{d} \Gamma '}{k_BT \cdot C(\theta)} \\
    & =- \frac{1}{k_BT} \int \nabla_\theta U_\theta(\Gamma ') P_\theta(\Gamma ') \dif \Gamma ' \\
    &= - \frac{1}{k_BT}  \EE_{\Gamma}[\nabla_\theta U_\theta(\Gamma)]\,.
    \end{aligned}
    $$
    
    Putting it all together, we have,
    
    $$
    \begin{aligned}
    \frac{\partial \EE_{\Gamma \sim P_\theta(\Gamma)}[g(\Gamma)]}{\partial \theta} & =  - \frac{1}{k_BT} \left ( \underbrace{\int g(\Gamma ') \nabla_\theta U_\theta(\Gamma ')^{\top} P_\theta(\Gamma ')\dif \Gamma '}_{\EE\left[g(\Gamma) \cdot \nabla_\theta U_\theta(\Gamma)\right]} -  \underbrace{ \left ( \int \cdot g(\Gamma ') \cdot  P_\theta(\Gamma ') \cdot \mathrm{d} \Gamma ' \right )}_{\EE[g(\Gamma)]} \EE[\nabla_\theta U_\theta(\Gamma)]^{\top} \right) \\
    & =  - \frac{1}{k_BT} \left ( \EE\left[g(\Gamma) \cdot \nabla_\theta U_\theta(\Gamma)^{\top}\right] - \EE[g(\Gamma)] \EE[\nabla_\theta U_\theta(\Gamma)]^{\top} \right ) \\ 
    & = -\frac{1}{k_BT}\operatorname{Cov}\left(g(\Gamma), \nabla_\theta U_\theta(\Gamma) \right ).
    \end{aligned}
    $$
    In this work, we use the following unbiased estimator for covariance. Given 2 random vectors $X,Y$, and samples of these vectors $\{(X_1, Y_1), ... (X_n, Y_n)\}$, our estimator is given by:
    
    $$
        \widehat{\operatorname{Cov}}(X,Y) = \frac{1}{N-1} \sum_{i=1}^{N} X_i {Y_i}^{\top} - \frac{1}{N(N-1)} \left(\sum_{j=1}^{N} X_j \right) \left( \sum_{k=1}^{N} {Y_k}^{\top} \right ).
    $$
    We can prove it is unbiased by taking the expectation of the right hand side and applying linearity of expectation: \\
    $$
    \begin{aligned}
         \add{\widehat{\operatorname{Cov}}(X,Y)} &= \frac{1}{N-1} \sum_{i=1}^{N} \EE[X_i {Y_i}^{\top}] - \frac{1}{N(N-1)} \left(\sum_{i=1}^{N} \EE[X_i {Y_i}^{\top}]  + \sum_{j \neq k} \EE[X_j {Y_k}^{\top}] \right ) \\
         &= \frac{N}{N-1} \EE[XY^{\top}] - \frac{1}{N(N-1)} \left (N \cdot \EE[XY^{\top}] + N(N-1) \cdot \EE[X]\EE[Y]^{\top} \right ) \\  \\
         &= \EE[XY^{\top}] - \EE[X]\EE[Y]^{\top} = \operatorname{Cov}(X, Y).
    \end{aligned}
    $$

    Given the samples $\{\Gamma_1, , ... \Gamma_N\}$, we can then use the estimator above to get,  
    $$
    \begin{aligned}
    \widehat{\operatorname{Cov}}\left(g(\Gamma), \nabla_\theta U_\theta(\Gamma) \right ) &= \frac{1}{N - 1} \sum_{i=1}^{N} g(\Gamma_i)  \nabla_\theta U_\theta(\Gamma_i)^{\top} -  \frac{1}{N(N-1)} \left(\sum_{j=1}^{N} g(\Gamma_j) \right)\left( \sum_{k=1}^{N} \nabla_\theta U_\theta(\Gamma_k)\right)^{\top} \\
    &= \frac{N}{N - 1} \left[ \hat{\EE}\left[{g(\Gamma)  \nabla_\theta U_\theta(\Gamma)^{\top}}\right] - \hat{\EE}\left[g(\Gamma)\right]\hat{\EE}\left[{\nabla_\theta U_\theta(\Gamma)}\right]^{\top}\right] .
    \end{aligned}
    $$
    \\
    
    The final Boltzmann estimator is thus given as,

    $$
    \mathcal{E}(\Gamma_1, \ldots, \Gamma_N) \defas \frac{N}{k_BT(N - 1)} \left[\hat{\EE}\left[g(\Gamma)\right]\hat{\EE}\left[{\nabla_\theta U_\theta(\Gamma)}\right]^{\top} - \hat{\EE}\left[{g(\Gamma) \nabla_\theta U_\theta(\Gamma)}^{\top}\right] \right].
    $$
\end{proof}
We note that a similar estimator is obtained in \citep{difftre} by differentiating through a reweighting scheme arising from thermodynamic perturbation theory. We have presented an alternative derivation that does not require reweighting.

\subsection{Extension to Other Statistical Ensembles} \label{sec: other_ensembles}

\add{The Boltzmann Estimator is applicable out-of-the-box to any statistical ensemble where the probability of a microstate can be written as $P_\theta(\Gamma) \propto \exp \left(- \frac{1}{k_BT} \left[\mathcal{H}_\theta(\Gamma) + \mathcal{X}(\Gamma)\right]\right)$, where $\mathcal{X}$ contains state-dependent thermodynamic variables. For the isothermal-isobaric (NPT) ensemble, $\mathcal{X}(\Gamma) = p V(\Gamma)$, where $p$ is the simulation pressure and $V(\Gamma)$ is the volume of the microstate. For the grand canonical ($\mu V T$) ensemble, $\mathcal{X}(\Gamma) = \mu N(\Gamma)$, where $\mu$ is the chemical potential and $N(\Gamma)$ is number of particles in the microstate. For the canonical (NVT) ensemble considered in this work, $\mathcal{X}(\Gamma) = 0$.  Since $\mathcal{X}(\Gamma)$ is independent of the MLFF parameters $\theta$ in all cases, it can effectively be absorbed into the kinetic energy component of the Hamiltonian, which does not affect the computation of the estimator. Our derivation thus proceeds in the same fashion and yields the same estimator. The Localized Boltzmann Estimator also holds as before.}

\subsection{StABlE Training Algorithm} \label{sec: algorithm}
We provide an algorithmic description of our StABlE-Training procedure. 
\begin{algorithm}
\caption{StABlE Training Procedure}\label{algo}
\begin{algorithmic}[1]
\State \textbf{Initialize:} \\
Pre-trained Machine Learning Force Field $U_{\theta}$ \\
Reference energy and forces dataset $\mathcal{D}_{\text{train}}$
and observables $\{g_{\text{ref}}^{(i)}\}_{i=1}^N$ \\
Simulation length $t$, 
number of parallel replicas $R$, 
minimum unstable threshold $f_{min}$,\\ 
maximum unstable threshold $f_{max}$, 
energy and forces loss weight $\lambda$, 
learning rate $\alpha$
\\
\State $\Bar{\Gamma}_{curr} \leftarrow \{\Gamma_1(0), \Gamma_2(0), \ldots \Gamma_R(0)\} \sim \mathcal{D}_{\text{train}}$
\State Current fraction of unstable replicas $f_{unst} \leftarrow 0$
\State Total simulated time $T_{f_i} \leftarrow 0,  \forall i = 1, \ldots, R$
\State Mark all replicas as active for simulation
\Repeat
  \begin{tikzpicture}[overlay]
\draw  [decoration={brace,amplitude=10pt},decorate,line width=1.5pt]
(10.5, 0.2) -- (10.5,-2.5) node [black,midway,xshift= 1.7cm] 
{\footnotesize \textbf{Simulation Phase }};
\end{tikzpicture}
 \While{$f_{unst} < f_{max}$}
    \State Simulate active replicas w/ $U_{\theta}$ for $t$ steps starting from $\Bar{\Gamma}_{curr}$
    \State $T_{f_i} \leftarrow T_{f_i} + t$,  $\forall i$  
     corresponding to stable replicas
    \State $\Bar{\Gamma}_{curr} \leftarrow \{\Gamma_1({T_{f_1}}), \Gamma_2({T_{f_2}}), \ldots \Gamma_R({T_{f_R}})\}$ 
    \State Update $f_{unst}$ and mark unstable replicas inactive
  \EndWhile
  \\

  \State // $\Bar{\Gamma}_{curr}$ now contains $\{\Gamma_1({T_{f_1}}), \Gamma_2({T_{f_2}}), \ldots \Gamma_R({T_{R}})\}$, where $T_{f_i}$'s are per-replica total simulation times. At least $f_{max}$ fraction of replicas are unstable at this point

    \begin{tikzpicture}[overlay]
\draw  [decoration={brace,amplitude=10pt},decorate,line width=1.5pt]
(14,-.3) -- (14,-4.3) node [black,midway,xshift=1.7cm] 
{\footnotesize \textbf{Learning Phase}};
\end{tikzpicture}
    \While{$f_{unst} > f_{min}$}
        \State Rewind all trajectories by $t$ timesteps: $\Bar{\Gamma}_{curr} \leftarrow \{\Gamma_1({T_{f_1} - t}), \ldots ,\Gamma_R({T_{f_R} - t})\}$ 
        \State Simulate all replicas w/ $U_{\theta}$ for $t$ steps starting from $\Bar{\Gamma}_{curr}$
        \State Update $f_{unst}$ 
        \State Compute observables $\{\widehat{\EE}_{\Gamma}[g^{(i)}(\Gamma)]\}_{i=1}^N$ over all length-$t$ trajectories
        \State $\mathcal{L}_{obs} \leftarrow \sum_{i=1}^N \|\widehat{\EE}_{\Gamma}[g^{(i)}(\Gamma)] - g_{\text{ref}}^{(i)}\|^2$
        \State $\mathcal{L}_{\text{QM}} \leftarrow$ energy and forces loss of $U_\theta$ on dataset $\mathcal{D}_{\text{train}}$
        \State $\theta \leftarrow \theta - \alpha \cdot \nabla_\theta (\mathcal{L}_{obs} + \lambda \mathcal{L}_{\text{QM}})$ // Compute Boltzmann Estimator     
        
    \EndWhile
\State Mark stable replicas active, unstable replicas inactive for next simulation phase
\State // At most $f_{min}$ fraction of replicas are unstable at this point
\Until{Convergence or maximum cycles reached}

\end{algorithmic}
\end{algorithm}
\newpage

\subsection{Observables} \label{sec: observables}
We provide definitions and details of the observables considered in this work.

The distribution of interatomic distances serves as a low-dimensional description of 3D structure. For a configuration $r' \in \mathbb{R}^{N \times 3}$, it is defined as,
$$
h(r) = \frac{1}{N (N - 1)} \sum_{i=1}^{N} \sum_{\substack{j \neq i}}^{N} \delta\left(r - \left\| {r_i}{'} - {r_j}{'} \right\|\right),
$$
where $\delta$ is the Dirac-Delta function.
Although the observable need not be differentiable with respect to $r$ for our Boltzmann learning framework, we compute a differentiable version of $h(r)$ via Gaussian smearing as in \citep{Wang_2023} to facilitate comparison with differentiable simulation methods. 

The radial distribution function (RDF) captures how density (relative to the bulk) varies as a function of distance from a reference particle, and thus characterizes the structural/thermodynamic properties of the system. The RDF is defined as,    
 $$
 RDF(r) = \frac{V}{N^2 4\pi r^2} h(r).
 $$
 where $V$ is the volume of the simulation domain, $N$ is the number of particles, $r$ is the radial distance from a reference particle, and $h(r)$ is a histogram of pairwise distances. As with $h(r)$, we use Gaussian smearing to make the RDF differentiable. 

The velocity autocorrelation function (VACF) is an important dynamical observable. Many fundamental properties, such as the diffusion coefficient and vibrational spectra, are functions of this observable. Computing the VACF requires a window of consecutive simulation states to compute. The VACF at a given time lag $\Delta t$ is given by,
$$
VACF(\Delta t) = \frac{1}{S} \sum_{t_0} \sum_i <v_i(t_0), v_i(t_0 + \Delta t)>,
$$
where $t_0$ is an initial time, $v_i(t)$ is the velocity of the $i^{th}$ atom at timestep $t$, $<\cdot, \cdot>$ is an inner product, and $S$ is the total number of samples considered given the summations over initial times and atoms. In this work, we compute the VACF over a window of 100 consecutive simulation timesteps, and normalize the values by the autocorrelation at $\Delta t = 0$ to restrict the range to \remove{$[0, 1]$} \add{$[-1, 1]$}.

The diffusivity coefficient is a fundamental transport property with crucial implications on the performance of energy storage systems, among other applications. Related to the time-derivative of the mean squared displacement, the diffusivity coefficient is defined as,
$$
D = \lim_{t \to \infty} \frac{1}{6t} \frac{1}{N} \sum_{i=1}^{N} \left| r_i(t) - r_i(0) \right|^2
$$,
where $r_i(t)$ is the coordinate of the $i^{th}$ particle at time $t$ and $N$ is the number of atoms considered. For the water system considered in this work, we measure the diffusivity of all 64 oxygen atoms.

\subsection{Stability Criteria} \label{sec: stability_criteria}
We provide definitions and details on the stability criteria considered in this work.

Adapted from \citep{fu2022forces}, the maximum bond length deviation metric captures unphysical bond stretching or collapse in small flexible molecules. According to this criterion, a simulation becomes unstable at time $T$ if,
$$\max_{(i,j) \in \mathcal{B}} \left| \left( \left\| r_i(T) - r_j(T) \right\| - b_{i,j} \right) \right| > \Delta,
$$
where $\mathcal{B}$ is the set of all bonds, $i, j$ are the two endpoint atoms of the bond, and $b_{i,j}$ is the equilibrium bond length computed from the reference simulation. Following \citep{fu2022forces}, we set $\Delta = 0.5 A$ for final stability evaluation. However, we adopt a more conservative value of $\Delta = 0.25 A$ during training in order to detect and correct instability earlier. We use this criterion for the MD17 and MD22 datasets. 
 
The minimum intermolecular distance metric is used to detect unphysical coordination structures or collisions between molecules in the water system. According to this criterion, a simulation becomes unstable at time $T$ if,
$$\min_{(i,j) \notin \mathcal{B}}  \left\| r_i(T) - r_j(T) \right\| < \Delta,
$$
where $\mathcal{B}$ is the set of all bonds, and $i, j$ are the endpoint indices of two non-bonded atoms. We set $\Delta = 1.2 A$ to detect instability during training.

The minimum intermolecular distance metric is appropriate at train-time to detect local instability early before it cascades to the rest of the system. However, it is too sensitive to use for evaluation, as realistic simulation can still be achieved for some time after the occurrence of a highly localized instability. Therefore, following \citep{fu2022forces}, we adopt an instability metric based on the radial distribution function, defined as, 
$$
\int_{r=0}^{\infty} \left\| RDF_{\text{ref}}(r) - \left\langle RDF^{t}(r) \right\rangle_{t=T}^{T + \tau} \right\| \, dr > \Delta,
$$

where $\langle \cdot \rangle$ is the averaging operator, $\tau$ is a short time window, and $\Delta$ is the stability threshold. We use $\tau = 10$ ps and $\Delta = 3.0$ for water. The stability criterion is triggered if any of the three element-conditioned water RDFs (H-H, O-O, or H-O) exceeds the threshold.

\subsection{Architecture and Training Details} \label{sec: training_details}
We provide details on the model architectures and training procedures used in this work. MD simulations and MLFFs are written in the PyTorch framework and are built upon the MDsim \citep{fu2022forces} and Atomic Simulation Environment \citep{ase} packages. All training is performed on a single NVIDIA A100 GPU.

Supplementary Table \ref{table:model-settings} provides details on the MLFF architectures. $r_{max}$ is the cutoff distance used to construct the radius graph. $l_{max}$ denotes the level of E(3) equivariance used in the network.
\begin{table}[H]
\centering
\begin{tabular}{|l|c|c|c|c|}
\hline
\textbf{} & \textbf{Symmetry} &
\textbf{Parameter} & 
\textbf{$r_{max}$} & 
$l_{max}$\\ 
& \textbf{Principle} & \textbf{Count} & \textbf{(A)} & \\ \hline
\textbf{SchNet \citep{schnet}} & E(3)-invariant & 0.12M & 5.0 & - \\ \hline
\textbf{NequIP \citep{nequip}} & E(3)-equivariant & 0.12M & 5.0 & 1\\ \hline
\textbf{GemNet-T \citep{gemnet}} & SE(3)-equivariant & 1.89M & 5.0 & -\\ \hline
\end{tabular}
\caption{MLFF Architecture Details.}
\label{table:model-settings}
\end{table}

For energy and forces pre-training, we follow the protocols in \citep{fu2022forces}. In order to isolate the effect of observable-based learning, we begin StABlE Training only after pre-training has fully converged (that is, when $\mathcal{L}_{\text{QM}}$ has reached a plateau). This means that any improvements in stability or accuracy as a result of StABlE Training can be attributed to the learning signal from the reference observable, as opposed to the regularization from the QM energy and forces data.

We include relevant settings used for StABlE Training in Supplementary Table \ref{table:training-settings}. $\alpha$ is the learning rate, $\lambda$ is the strength of energy and forces regularization, $t$ is the number of simulation timesteps per epoch, and $R$ is the number of parallel replicas. 
We note that in practice, we compute the outer products and empirical means in the Boltzmann estimator in batched fashion. Thus, to limit memory usage, we compute $\frac{N}{B}$ separate Boltzmann estimators from minibatches of $B < N$ states and subsequently average them to produce a final estimator.

\addblue{\paragraph{General Guidelines for Choosing Hyperparameters.} $t$ should be chosen large enough that the deviation of ensemble averages computed within the window from ground truth values are primarily attributable to systematic error/physical instability rather than sampling error. If t is chosen too large, the frequency of gradient updates reduces, slowing down learning. Generally, a frequency of 1 picosecond should be sufficient for structural observables of small molecular systems, and may need to be larger (10-100 ps) for larger-scale or coarse-grained systems. The number of replicas $R$ should be chosen so as to maximize MLFF inference throughput (samples/second) while remaining within GPU memory. Since we perform simulations in parallel by vectorizing over the batch dimension, we see steady improvements in throughput until GPU memory saturates, at which point performance plateaus or degrades. The minibatch size $B$ should also be chosen as large as possible to minimize variance in the Boltzmann Estimator, while remaining within GPU memory limits.}

\begin{table}[H]
\centering
\small
\begin{tabular}{ |c|c|c|c|c|c|c|c|c|c| }
 \hline
 \textbf{} & \textbf{MLFF} & \textbf{Stability} & \textbf{Training} & \textbf{Estimator} & \textbf{\(\alpha\)} & \textbf{\(\lambda\)} & \textbf{t} & $R$ & $B$ \\
 & & \textbf{Criterion, Threshold} & \textbf{Observable} & \textbf{Type} & & & & & \\
 \hline
 \textbf{Aspirin} & SchNet & Bond Len. Dev., 0.25 A & $h(r)$ & Global & 0.001 & 10 & 2000 & 128 & 40 \\ \hline
 \textbf{Ac-Ala3-NHMe} & NequIP & Bond Len. Dev., 0.25 A & $h(r)$ & Global & 0.001 & 10 & 2000 & 128 & 40 \\ \hline
 \textbf{Water} & GemNet-T & IMD, 1.2 A & O-H Bond Length & Local & 0.003 & 0 & 1000 & 8 & 4 \\ \hline
\end{tabular}
\caption{StABlE Training Settings.}
\label{table:training-settings}
\end{table}

\paragraph{Wall Clock Time of StABlE Training. } In Supplementary Table \ref{table:training-time}, we provide the total wall clock time spent on QM pre-training, as well as the subsequent StABlE Training. All runtimes were measured on an NVIDIA A100 GPU. We note that especially for Ac-Ala3-NHMe and Water, StABlE Training incurs a relatively small marginal computational cost beyond that of QM pre-training.
\begin{table}[H]
\centering
\begin{tabular}{|l|c|c|}
\hline
\textbf{} & \textbf{QM pre-training} & \textbf{StABlE Training} \\ \hline
\textbf{Aspirin} & 2 & 4 \\ \hline
\textbf{Ac-Ala3-NHMe} & 16 & 4.7 \\ \hline
\textbf{Water} & 31 & 3.5 \\ \hline
\end{tabular}
\caption{Wall clock time, in hours, of QM pre-training and StABlE Training.}
\label{table:training-time}
\end{table}

\subsection{Simulation Details} \label{sec: simulation_details}
We provide MD simulation details in Supplementary Table \ref{table:simulation-parameters}. \addtwo{During training,} all systems \removetwo{considered in this work} are simulated with a Nose-Hoover thermostat. \addtwo{During evaluation, either a Nose-Hoover or Langevin thermostat is used based on whichever one yields better stability}. Thermostat parameters are chosen to be consistent with prior literature \citep{md17, md22, fu2022forces}: for Nose-Hoover simulations, the temperature coupling constant is set to 20 fs, and for Langevin simulations, the friction coefficient is set to 0.1 ps$^{-1}$. 
\begin{table}[H]
\centering
\begin{tabular}{|l|c|c|c|c|}
\hline
\textbf{} & 
\textbf{Temperature} & 
\textbf{Timestep} & 
\textbf{Periodic Boundary} & 
\textbf{Simulation} \\ 
& \textbf{(K)} & \textbf{(fs)} & \textbf{Conditions} & \textbf{Thermostat} \\ \hline
\textbf{Aspirin} & 500 & 0.5 & No & Langevin\\ \hline
\textbf{Ac-Ala3-NHMe} & 500 & 0.5 & No & Nose-Hoover\\ \hline
\textbf{Water} & 300 & 1 & Yes & Nose-Hoover\\ \hline
\end{tabular}
\caption{Simulation Settings.}
\label{table:simulation-parameters}
\end{table}

\addtwo{For simulating water in the NPT ensemble at 300K and 1 atm, we employ the Berendsen barostat with temperature and pressure coupling times of 20 fs and 2 ps respectively. We limit the per-timestep change in momentum and volume to 10\% and 1\% respectively to prevent instabilities resulting from large fluctuations. We found the Berendsen barostat to be more stable than combined Parinello-Rahman and Nose-Hoover dynamics, which would be the more conventional choice for NPT simulations with standard potentials. We speculate that this may be due to qualitative differences in the behavior of ML potentials compared to classical or \textit{ab-initio} potentials, and these differences warrant further investigation in the future.
}

\subsection{Evaluation Details} \label{sec: evaluation_details}
We provide further details on the protocol used to evaluate the MLFFs considered in this work. 
Our evaluation protocol is centered around MD simulations. To facilitate direct comparison between StABlE-trained MLFFs and those trained only on energy and forces reference data, for a given molecular system we perform MD simulations starting from the same initial configurations for all models. We choose the number of parallel replicas and total simulation time on a per-system basis so as to saturate GPU memory usage while remaining within a reasonable computational budget. Simulation conditions are the same as described in Section \ref{sec: simulation_details} except for the temperature generalization experiment, in which the temperature of simulation is varied. Supplementary Table \ref{table:evaluation-parameters} summarizes the relevant evaluation parameters for each system.

\begin{table}[H]
\centering
\begin{tabular}{|l|c|c|c|}
\hline
\textbf{} & 
\textbf{Num. Parallel Replicas} & 
\textbf{Max. Simulation Time (ps)} & 
\textbf{Stability Criterion, Threshold} \\ \hline
\textbf{Aspirin} & 256 & 1000 & Bond Length Deviation, 0.5 A \\ \hline
\textbf{Ac-Ala3-NHMe} & 48 & 300 & Bond Length Deviation, 0.5 A \\ \hline
\textbf{Water} & 5 & 1000 & RDF MAE, 3.0 \\ \hline
\end{tabular}
\caption{StABlE Evaluation Settings.}
\label{table:evaluation-parameters}
\end{table}

\begin{figure}[H]
    \centering
    \includegraphics[width=1.0\textwidth] {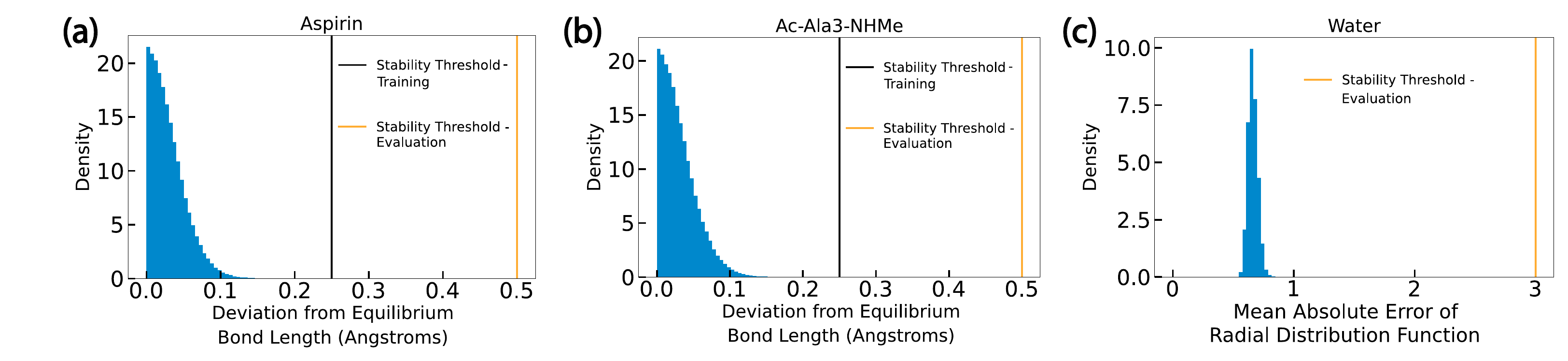}
    \caption{\textbf{Distribution of stability criterion over reference simulations.} Instability thresholds are chosen to be very relaxed, such that crossing of the threshold signifies catastrophic, unrecoverable instability.}
    \label{fig:stability_thresholds}
\end{figure}

\paragraph {Justification of Chosen Stability Thresholds.}
Following \citep{fu2022forces}, we choose stability thresholds such that a realistic, high-fidelity simulation at the chosen temperature would virtually never cross the threshold. This means that if a simulation does cross the threshold, this is indicative of catastrophic failure. Thresholds are set more conservatively during training in order to facilitate early detection of potential collapse. Supplementary Figure \ref{fig:stability_thresholds} shows the distribution of values of the stability criterion over high-fidelity reference simulations for the three systems considered in this work, along with thresholds chosen to denote instability for training and evaluation. \addblue{As a rough guideline for new systems, we suggest setting the threshold at 4 standard deviations beyond mean fluctuations for training, and 5 standard deviations for evaluation.}

\subsection{Temperature-Reweighting of Observables} \label{sec: temperature_reweighting}

\begin{figure}
    \centering
    \includegraphics[width=0.8\textwidth] {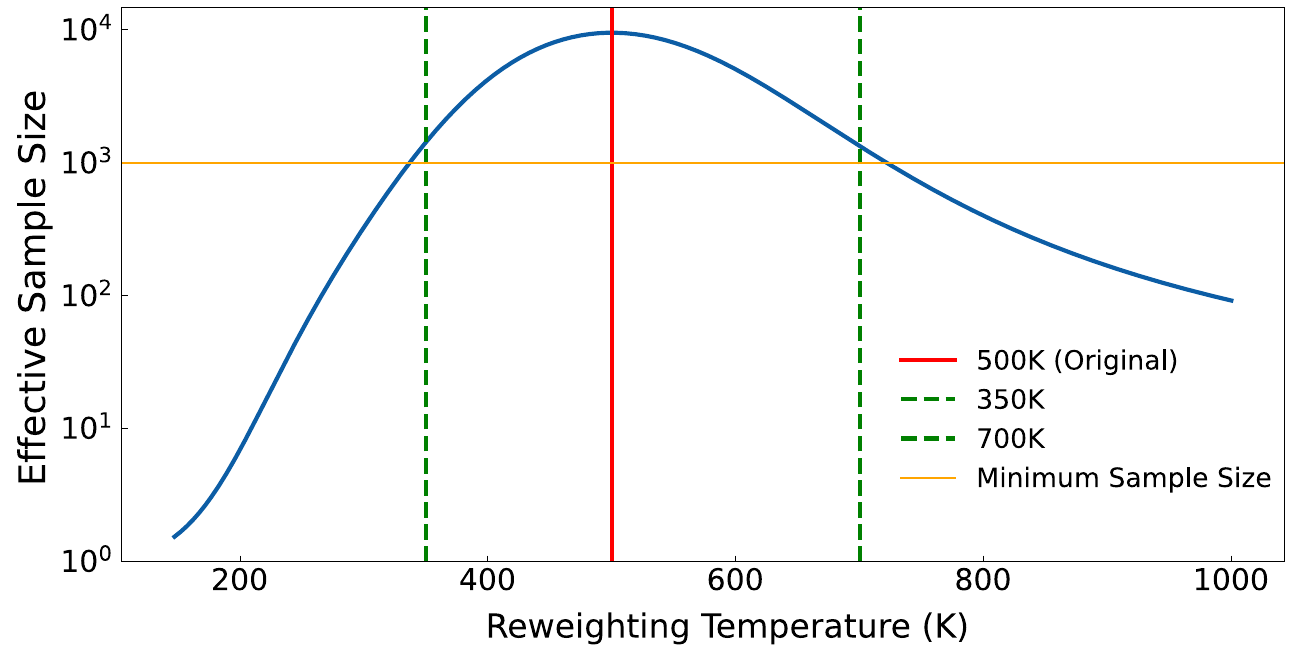}
    \caption{\textbf{Boltzmann-reweighting of aspirin samples.} The effective sample size ($N_{eff}$) as a function of reweighting temperature for aspirin dataset. $N_{eff}$ is maximized when the reweighting temperature is equal to the original temperature (500K). Using a minimum sample size of 1000, we choose upper and lower temperatures of 700K and 350K at which to perform temperature generalization experiments.}
    \label{fig:reweighting}
\end{figure}
We provide further details on the reweighting process used to estimate the reference distribution of interatomic distances at 350K and 700K. This was used in the temperature generalization experiments described in Section \ref{sec:aspirin_temperature}.

Under the canonical ensemble, microstates follow a Boltzmann distribution $P_\theta(\Gamma) \defas \frac{\exp \left(- \frac{1}{k_BT} \mathcal{H}_\theta(\Gamma)\right)}{C(\theta)}$. Consider states $\Gamma_1, \ldots, \Gamma_N$ sampled at temperature $T_1$. Define a reweighting factor for each sample as follows,
$$
\begin{aligned}
w_i &= \frac{\frac{P_\theta(\Gamma; T_1)}{P_\theta(\Gamma; T_2)}}{\sum_{i=1}^N \frac{P_\theta(\Gamma_i; T_1)}{P_\theta(\Gamma_i; T_2)}}
&= \frac{\exp(-\frac{\mathcal{H}_\theta(\Gamma_i)}{k_B} (\frac{1}{T_2} - \frac{1}{T_1}))}{\sum_{i=1}^N \exp(-\frac{\mathcal{H}_\theta(\Gamma_i)}{k_B} (\frac{1}{T_2} - \frac{1}{T_1}))}.
\end{aligned}
$$

We can then compute a reweighted Monte Carlo estimate of the observable at $T_2$ as follows \citep{difftre}.
$$g_{true, T_2} = \sum_{i=1}^N w_i g(\Gamma_{(i)})$$

The statistical error of the reweighted Monte Carlo estimate is captured by the effective sample size, 
$N_{eff} \approx e^{-\sum_{i=1}^N w_i ln(w_i)}$ \citep{Carmichael2012Multiscale}. A small effective sample size indicates that a few samples with high weights dominate the average; this occurs for large differences between $T_1$ and $T_2$. To select lower and upper temperatures at which to perform the temperature generalization experiment, we set a minimum $N_{eff} = 1000$, leading us to choose 350K and 700K (Figure \ref{fig:reweighting}).

\vspace{-20pt}
\subsection{StABlE Training in the Isothermal-Isobaric Ensemble} \label{sec:npt}

We repeat StABlE Training on the all-atom water system, and this time simulate in the isothermal-isobaric (NPT) ensemble with a temperature of 300K and a pressure of 1 atm. We use the same training settings outlined in Supplementary Table \ref{table:training-settings}. We find similar results to when we simulate in the canonical (NVT) ensemble: StABlE Training yields clear stability improvements, increasing the median stable simulation time from 51 to 165 picoseconds. The stability and quality of estimated observables is slightly lower than in NVT simulations, including some unphysical collisions in the short-range region of the element-conditioned RDFs. This may be due to the distribution shift induced by the continuously changing box size in NPT simulations, which was not seen during pretraining of the GemNet-T potential.

\begin{figure}[H]
    \centering
    \includegraphics[width=1.0\textwidth] {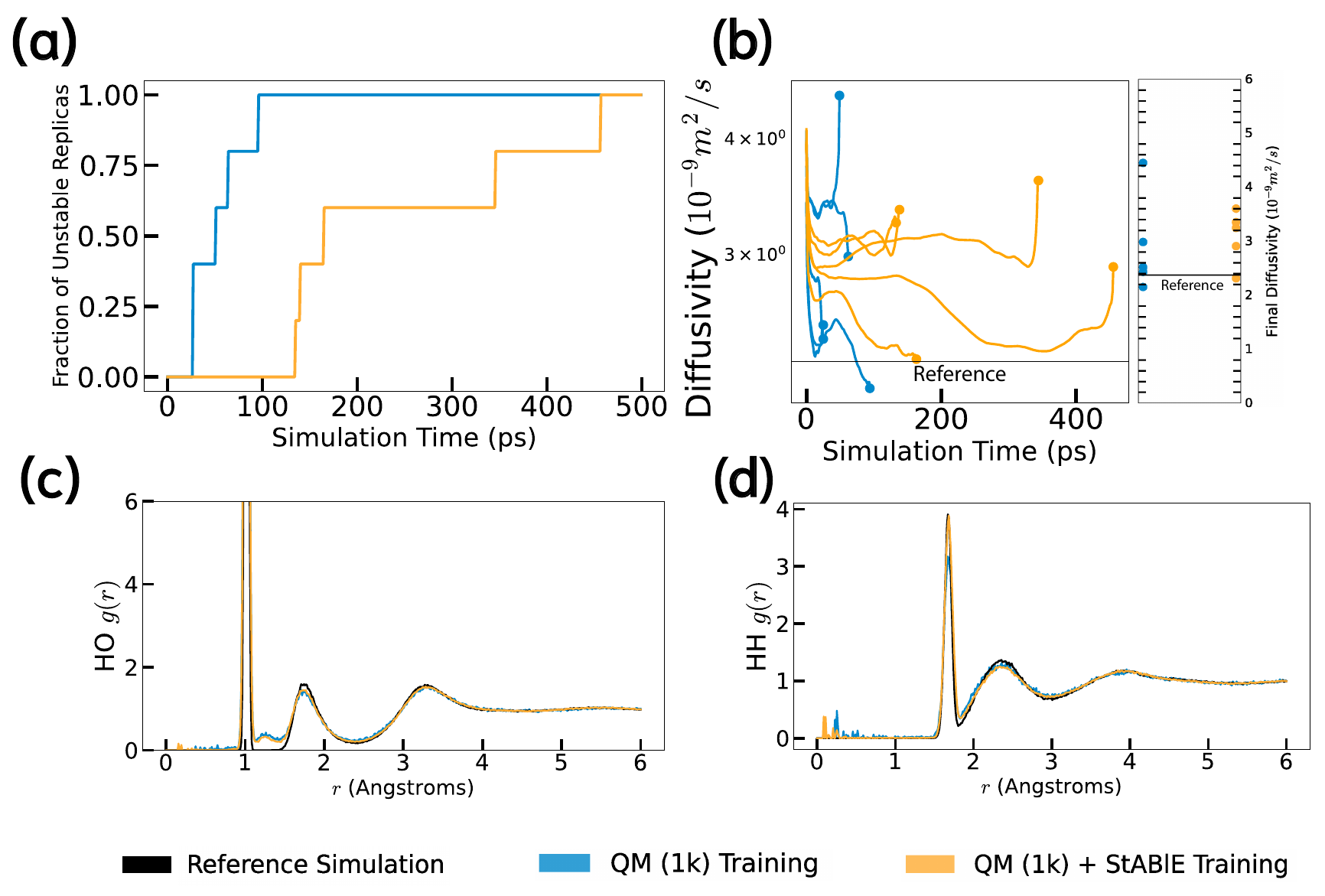}
    \caption{\textbf{Results of StABlE Training a GemNet-T model for all-atom water simulation in the isothermal-isobaric (NPT) ensemble.} StABlE Training yields considerable improvements in stable simulation time, and maintains or slightly improves the accuracy of recovered observables.}
    \label{fig:npt}
\end{figure}

\subsection{Analysis of Energy and Forces Errors} \label{sec:energy_force_analysis}
We study the effect of two hyperparameters of StABlE Training, the learning rate $\alpha$ and the strength of QM regularization $\lambda$, on the energy and forces errors of a SchNet MLFF on a held-out test set of aspirin structures. We perform StABlE Training for learning rates ranging from $10^{-5}$ to $10^{-3}$ and QM regularization coefficients ranging from $10^0$ to $10^2$. We perform evaluation of each trained model via MD simulation of 256 parallel replicas at 500K. 
\begin{figure}[H]
    \centering
    \includegraphics[width=1.0\textwidth] {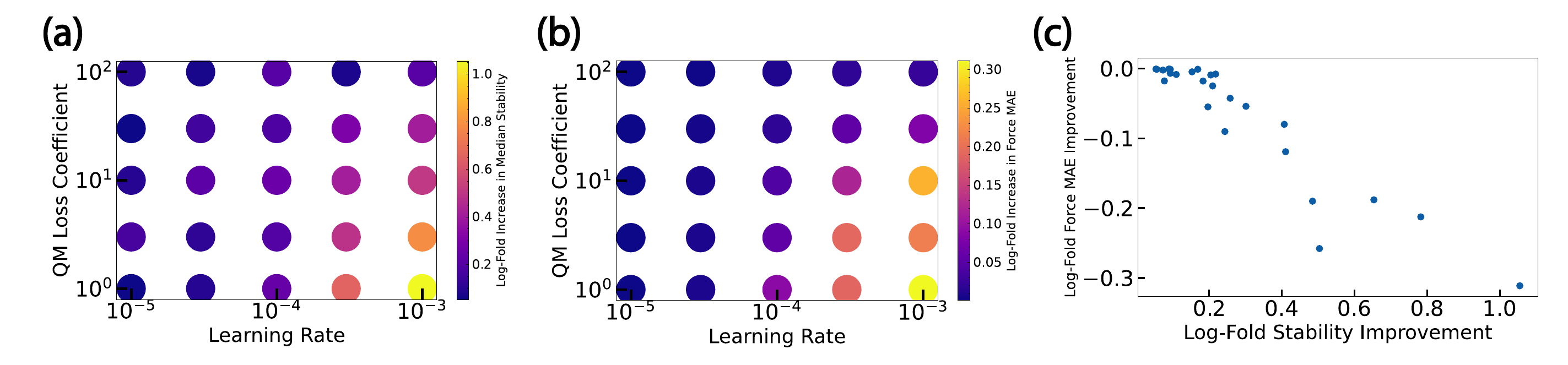}
    \caption{\textbf{Effect of training hyperparameters on stability and force error improvements.} \textbf{(a)} Models trained with higher learning rate and lower QM loss coefficient achieve better stability gains relative to a baseline model trained only on QM reference data. \textbf{(b)} Models trained with higher learning rate and lower QM loss coefficient incur higher increases in force mean absolute error (MAE) on a held-out test dataset relative to the baseline model. \textbf{(c)} A Pareto frontier of Stability vs Force MAE arises. Some choices of learning rate and loss coefficient are Pareto-suboptimal, while choosing others moves along the Pareto frontier.}
    \label{fig:energy_force_analysis}
\end{figure}

Training runs with high learning rate and low QM loss coefficient achieve greater improvements in stable simulation time relative to a baseline model trained only on QM reference data (Supplementary Figure \ref{fig:energy_force_analysis}a). However, these training runs also incur a greater increase in the Mean Absolute Error (MAE) of force prediction on a held out test dataset (Supplementary Figure \ref{fig:energy_force_analysis}b). Due to training on a single structural observable, the observable-matching component of the StABlE objective is ill-posed: a MLFF which collapses simulations onto a sparse set of states exactly matching the reference observable would globally minimize the observable-matching loss and yield indefinitely stable simulations, while incurring a large QM/force error. As the learning rate of the StABlE Training procedure is increased, the optimization is increasingly pushed towards this degenerate mode. Increasing the weight of the QM objective counteracts this tendency. Consequently, a Pareto frontier arises between stability and force prediction accuracy (Supplementary Figure \ref{fig:energy_force_analysis}c). Some settings of learning rate and QM loss coefficient are Pareto suboptimal, while choosing among the remaining combinations causes one to move along the frontier. Incorporating additional training observables, particularly those which are dynamical in nature (e.g., velocity autocorrelation functions), could counteract the degeneracy and push the Pareto frontier outwards. 

Finally, we note that the observed energy MAE increase on aspirin for our chosen combination of learning rate ($\alpha = 0.001$) and loss coefficient ($\lambda = 10$) is from 0.87 to 1.4 $\mathrm{kcal\,mol}^{-1}$, while DFT error for energies on MD17 can be as high as 2.3 $\mathrm{kcal\,mol}^{-1}$ \citep{Faber2017PredictionErrors}. Thus, some of the error in the MLFF predictions could be attributable to inaccuracies in the underlying DFT data.

\addblue{As rough guidelines for new systems, if Force MAE is prioritized, then the learning rate should be smaller and the QM loss coefficient should be set higher. If stability improvements are prioritized over Force MAE, such as in cases where the reference energy/force data is known to be unreliable, the opposite is true.}

\subsection{Effect of Simulation Timestep on Stability} \label{sec:dt_analysis}

\begin{figure}[H]
    \centering
    \includegraphics[width=1.0\textwidth] {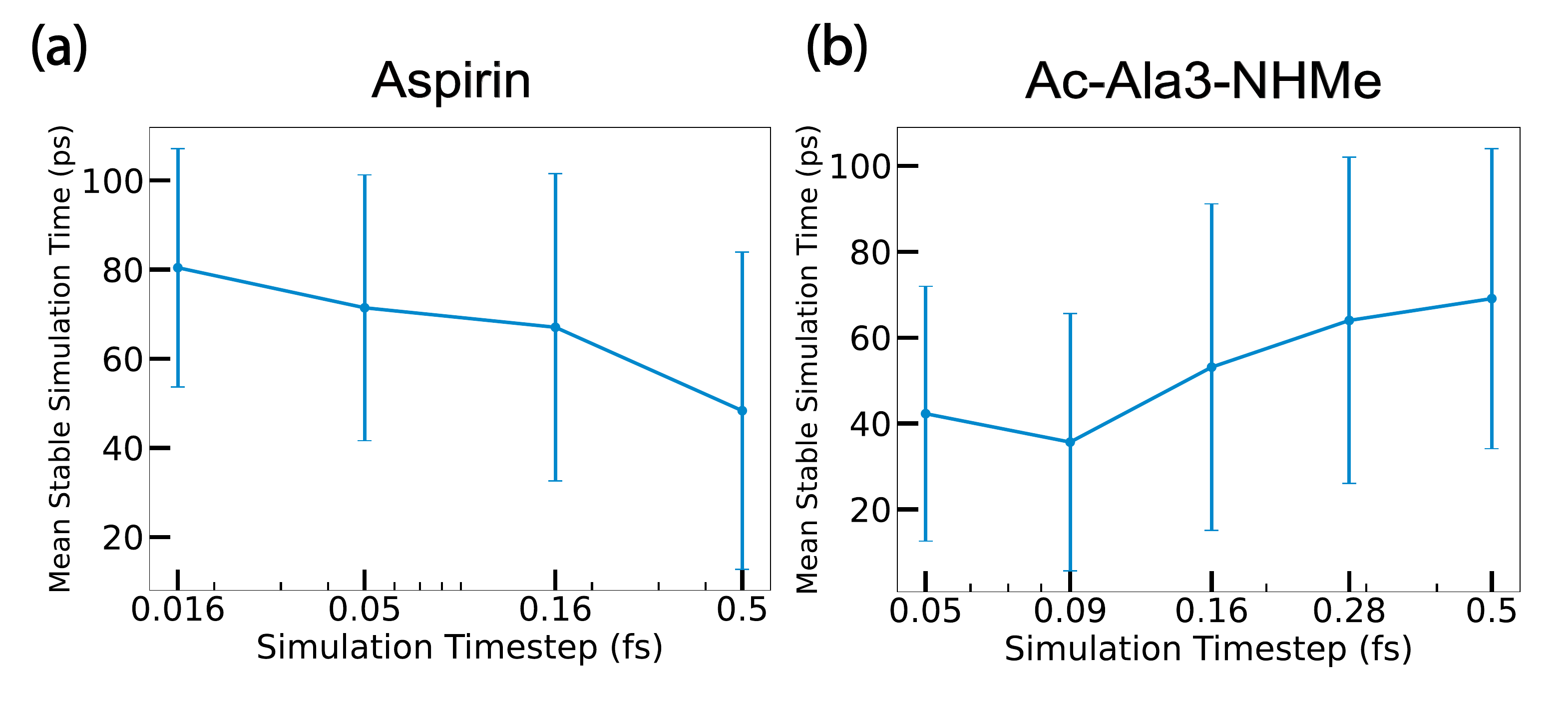}
    \caption{\textbf{Effect of reducing timestep on simulation stability.} Reducing the timestep does not completely eliminate instability, and can sometimes worsen stability.}
    \label{fig:dt_figure}
\end{figure}

We perform 100 ps simulations with 32 replicates, using various timesteps for the aspirin and Ac-Ala3-NHMe tetrapeptide systems, using a SchNet and NequIP potential respectively (Supplementary Figure \ref{fig:dt_figure}).

We observe that instability is not completely eliminated as the timestep is reduced. For the tetrapeptide system, stability consistently decreases as the timestep is reduced. Similar behavior has been observed in neural network based solvers for ordinary differential equations \citep{krishnapriyan2023learning}. For the aspirin system, stability improves as the timestep is reduced, but does so very slowly (simulations are not completely stable even with a timestep of 0.05 fs , which is 10 times lower than the original timestep).

\addblue{We also investigate the effect of increasing beyond the original timestep of 0.5 fs on simulation stability for Aspirin and Ac-Ala3-NHMe. We again perform 100 ps simulations with 32 independent replicates, now with timesteps of 1, 2, 5, and 10 fs. We observe that StABlE Training yields stability improvements at larger timesteps up to 2 fs, but after this point, neither the pretrained nor StABlE-trained potential are able to simulate stably for an appreciable amount of time (Supplementary Figure \ref{fig:longdt_figure}). We emphasize that we cutoff the simulations at 100 ps, so the aspirin simulation with the StABlE-trained model using a timestep of 1 fs would likely simulate stably for considerably longer if not cut off.

\begin{figure}[H]
    \centering
    \includegraphics[width=1.0\textwidth] {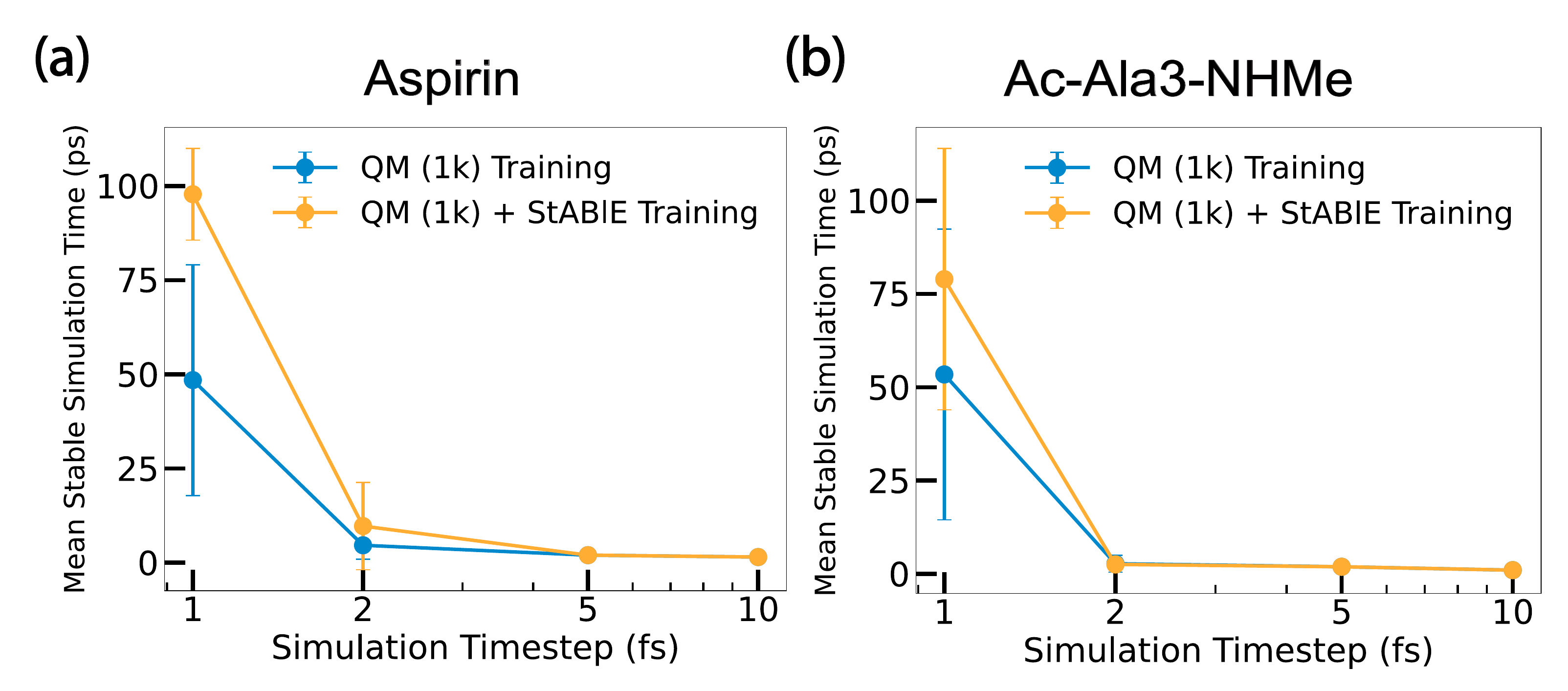}
    \caption{\textbf{Effect of increasing timestep on simulation stability.} StABlE Training improves simulation stability for timesteps up to 2 fs, after which stability rapidly deteriorates.}
    \label{fig:longdt_figure}
\end{figure}
}

\addblue{\subsection{Comparison of Boltzmann Estimator to Alternative Differentiation Strategies}
\label{sec:unrolled_comp}

We compare our Boltzmann estimator with two alternative differentiation strategies, namely direct backpropagation through the unrolled MD simulation, and the adjoint method described in \citep{chen2019neural}. As in \citep{Wang_2023}, we consider a system with 32 particles governed by a Lennard-Jones potential acting on the pairwise particle distances. We initialize the simulations with a prior potential capturing only the repulsive term of the potential, and seek to learn a correction term, parameterized by a multi-layer perceptron with 5 hidden layers of size 128, so as to reproduce the behavior of the full potential. We utilize supervision from the ground truth radial distribution function. We measure the loss gradient norms, memory footprint, and runtime of all approaches as a function of the simulation length, showing results in Figure \ref{fig:unrolled_fig}. As expected, direct backpropagation quickly runs out of memory because it needs to store intermediate network activations after every forward pass. The adjoint method eliminates this memory requirement by performing a backwards ODE solve to calculate the loss gradients. However, as reported in \citep{Wang_2023}, the adjoint dynamics are highly unstable over long rollouts and lead to exploding gradient norms. Meanwhile, our Boltzmann estimator achieves roughly constant gradient norms as the simulation length increases due to the decoupling of the gradient computation from the dynamics, and also has a favorably low memory and compute footprint.

\begin{figure}[H]
    \centering
    \includegraphics[width=1.0\textwidth] {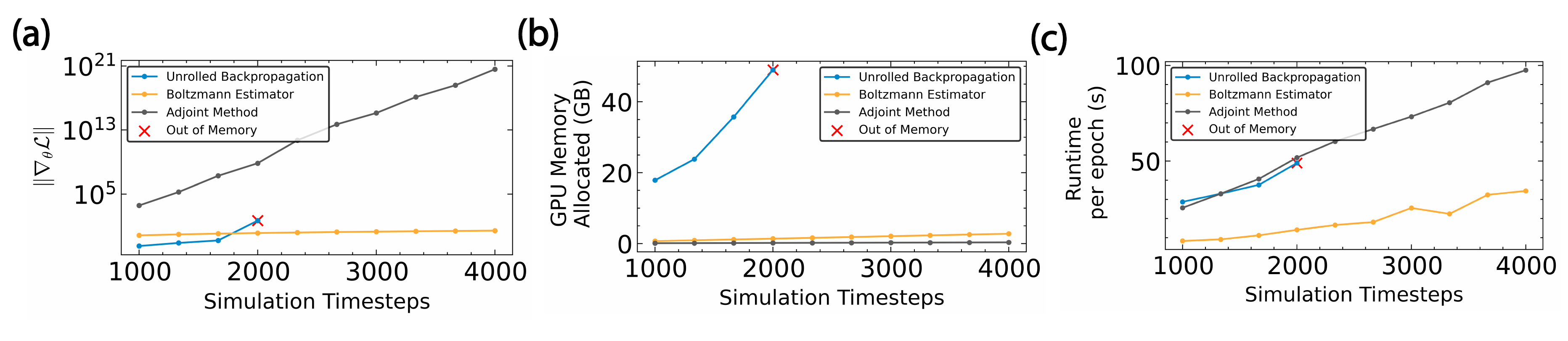}
    \caption{\addblue{\textbf{Comparison of Boltzmann Estimator to direct backpropagation and adjoint method on a toy Lennard-Jones system .} The Boltzmann Estimator achieves stable gradient norms and favorable memory and runtime footprints as the simulation length is increased. Direct backpropagation is memory prohibitive, and the adjoint method suffers from unstable dynamics, eventually causing gradient norms to explode.}}
    \label{fig:unrolled_fig}
\end{figure}
}

\subsection{Velocity Autocorrelation Function of Aspirin} \label{sec:vacf}

We show the aspirin velocity autocorrelation function (VACF) corresponding to the trajectory with the median stability improvement between conventional and StABlE Training. A StABlE-trained model produces a similar VACF relative to a conventionally trained model. This suggests that the StABlE procedure does not significantly interfere with dynamic properties of the simulation, despite only training with a structural observable ($h(r)$). 

\begin{figure}[H]
    \centering
    \includegraphics[width=0.7\textwidth] {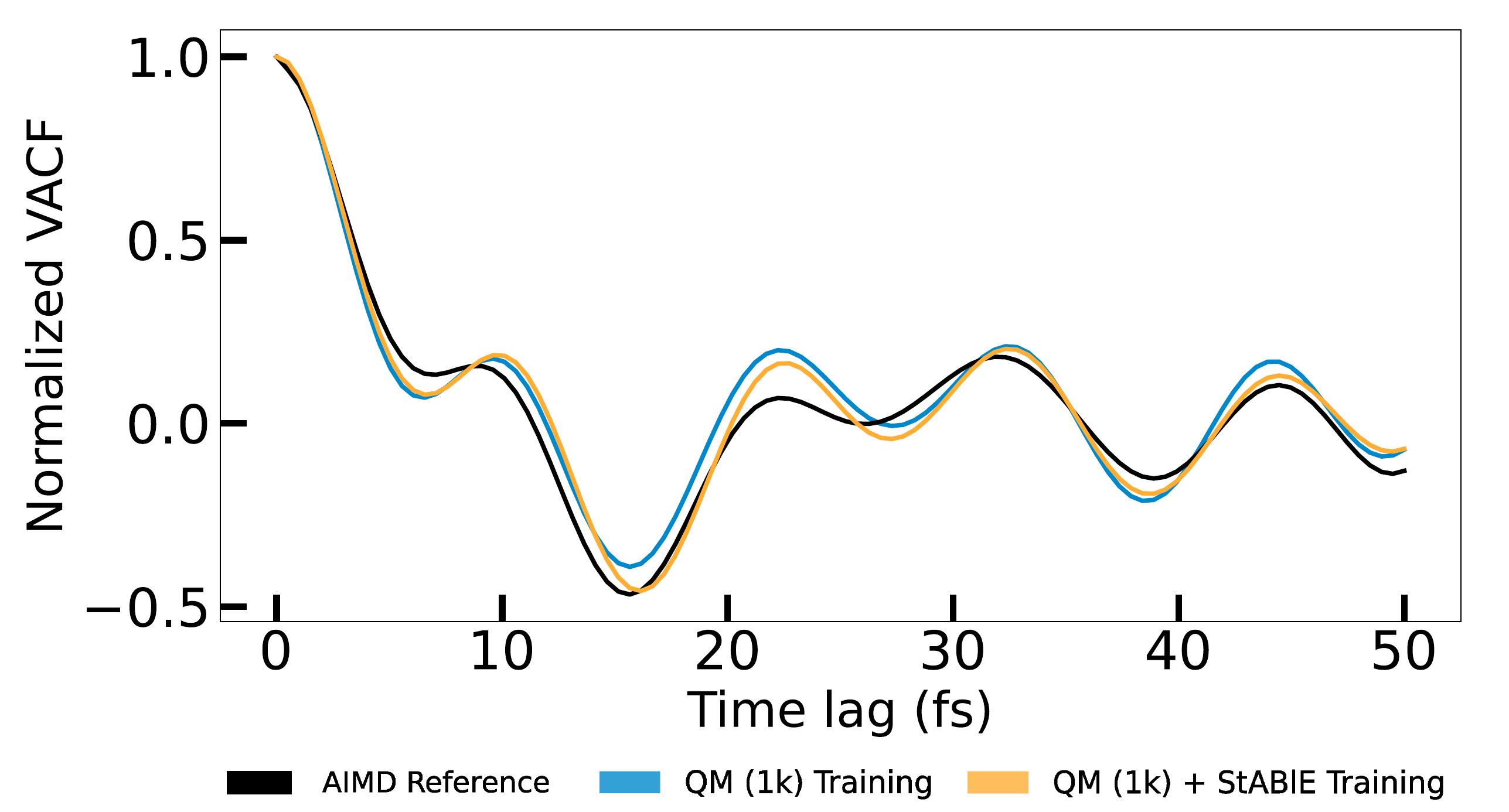}
    \caption{\textbf{Velocity autocorrelation function of aspirin.}}
    \label{fig:vacf_figure}
\end{figure}

\end{document}